\documentclass{article}

\usepackage[nonatbib,final]{neurips_2020}

\usepackage[utf8]{inputenc} % allow utf-8 input
\usepackage[T1]{fontenc}    % use 8-bit T1 fonts
\usepackage{hyperref}       % hyperlinks
\usepackage{url}            % simple URL typesetting
\usepackage{booktabs}       % professional-quality tables
\usepackage{amsfonts}       % blackboard math symbols
\usepackage{nicefrac}       % compact symbols for 1/2, etc.
\usepackage{microtype}      % microtypography

\usepackage{amsmath,amsfonts,bm}

\def\eqref#1{equation~\ref{#1}}
\def\1{\bm{1}}

\def\vb{{\bm{b}}}

\def\ve{{\bm{e}}}

\def\vh{{\bm{h}}}

\def\vk{{\bm{k}}}

\def\vm{{\bm{m}}}

\def\vq{{\bm{q}}}

\def\vw{{\bm{w}}}
\def\vx{{\bm{x}}}
\def\vy{{\bm{y}}}

\def\mA{{\bm{A}}}
\def\mB{{\bm{B}}}
\def\mC{{\bm{C}}}
\def\mD{{\bm{D}}}
\def\mE{{\bm{E}}}

\def\mK{{\bm{K}}}

\def\mM{{\bm{M}}}

\def\mP{{\bm{P}}}

\def\mU{{\bm{U}}}

\def\mW{{\bm{W}}}

\DeclareMathAlphabet{\mathsfit}{\encodingdefault}{\sfdefault}{m}{sl}
\SetMathAlphabet{\mathsfit}{bold}{\encodingdefault}{\sfdefault}{bx}{n}
\newcommand{\tens}[1]{\bm{\mathsfit{#1}}}

\def\tH{{\tens{H}}}

\def\tU{{\tens{U}}}

\def\sI{{\mathbb{I}}}

\def\sS{{\mathbb{S}}}

\newcommand{\R}{\mathbb{R}}

\usepackage{graphicx}
\usepackage{wrapfig}
\usepackage{amsmath}
\usepackage{amssymb}
\usepackage{amsthm}
\usepackage{subcaption}
\usepackage{tabularx}
\usepackage[linesnumbered,ruled,vlined]{algorithm2e}   % algorithm environment
\usepackage{biblatex}
\newcommand{\var}{\textbf{\textit{V}}}
\newcommand{\vard}[1]{\var\textbf{:#1}}
\newcommand{\nll}{\mathcal{L}_{\text{nll}}}
\theoremstyle{definition}

\newtheorem{step}{Step}

\makeatletter
\newcommand*{\inlineequation}[2][]{%
  \begingroup
    \refstepcounter{equation}%
    \ifx\\#1\\%
    \else
      \label{#1}%
    \fi
    \relpenalty=10000 %
    \binoppenalty=10000 %
    \ensuremath{%
      #2%
    }%
    ~\@eqnnum
  \endgroup
}
\makeatother

\usepackage[firstpage,nostamp]{draftwatermark}
\SetWatermarkColor[gray]{0.9}

\title{Learning Invariants through Soft Unification}

\author{%
  Nuri Cingillioglu\\
  Imperial College London\\
  \texttt{nuric@imperial.ac.uk}\\
  \And
  Alessandra Russo\\
  Imperial College London\\
  \texttt{a.russo@imperial.ac.uk}\\
}

\begin{document}

\maketitle

\begin{abstract}
Human reasoning involves recognising common underlying principles across many examples. The by-products of such reasoning are invariants that capture patterns such as ``if someone went somewhere then they are there'', expressed using variables ``someone'' and ``somewhere'' instead of mentioning specific people or places. Humans learn what variables are and how to use them at a young age. This paper explores whether machines can also learn and use variables solely from examples without requiring human pre-engineering. We propose Unification Networks, an end-to-end differentiable neural network approach capable of lifting examples into invariants and using those invariants to solve a given task. The core characteristic of our architecture is soft unification between examples that enables the network to generalise parts of the input into variables, thereby learning invariants. We evaluate our approach on five datasets to demonstrate that learning invariants captures patterns in the data and can improve performance over baselines.
\end{abstract}

\section{Introduction}

Humans have the ability to process symbolic knowledge and maintain symbolic thought~\cite{coevoflangandbrain}. When reasoning, humans do not require combinatorial enumeration of examples but instead utilise invariant patterns where specific entities are replaced with placeholders. Symbolic cognitive models~\cite{cogmodelsymbolic} embrace this perspective with the human mind seen as an information processing system operating on formal symbols such as reading a stream of tokens in natural language. The language of thought hypothesis~\cite{languageofthought} frames human thought as a structural construct with varying sub-components such as ``X went to Y''. By recognising what varies across examples, humans are capable of lifting examples into invariant principles that account for other instances.
%plausible biological neural models proposed such as for handling triplets of symbols in question answering~\cite{brainaiqa}.
This symbolic thought with variables is learned at a young age through symbolic play~\cite{piagetsymthought}. For instance, a child learns that a sword can be substituted with a stick~\cite{sticksword} and engages in pretend play.% Thus, we ask the question: if and how can a machine also learn and use the idea that a symbol can take on different values? For example, learning that the token \texttt{mary} could be \emph{someone} else and that person could go \emph{somewhere} else.

\setlength{\intextsep}{0.5em}
\begin{wrapfigure}{r}{0.5\linewidth}
\centering
\renewcommand{\arraystretch}{1.1}
\begin{tabular}{l}
  \vard{bernhard} is a \vard{frog}\\
  \vard{lily} is a \vard{frog}\\
  \vard{lily} is \vard{green}\\
  \hline
  what colour is \vard{bernhard}\\
  \hline
  green
\end{tabular}
\caption{Invariant learned for bAbI task 16, basic induction, where \vard{bernhard} denotes a variable with default symbol bernhard. This single invariant accounts for all the training, validation and test examples of this task.}
\label{fig:rule_babi_task16}
\end{wrapfigure}

Although variables are inherent in symbolic formalisms and their models of computation, as in first-order logic~\cite{russell2016artificial}, they are pre-engineered and used to solve specific tasks by means of assigning values to them. However, when learning from data only, being able to recognise when and which symbols could act as variables and therefore take on different values is crucial for lifting examples into general principles that are invariant across multiple instances. Figure~\ref{fig:rule_babi_task16} shows an example invariant learned by our approach: \emph{if someone is the same thing as someone else then they have the same colour}. With this invariant, our approach solves \emph{all} training and test examples in task 16 of the bAbI dataset~\cite{babi}.

In this paper we address the question of whether a machine can learn and use the notion of a \emph{variable}, i.e. a symbol that can take on different values. For instance, given an example of the form ``bernhard is a frog'' the machine would learn that the token ``bernhard'' could be \emph{someone} else and the token ``frog'' could be \emph{something} else.
When the machine learns that a symbol is a variable, assigning a value can be reframed as attending to an appropriate symbol.
Attention models~\cite{bahdanauatt, nmtatt, attsurvey} allow neural networks to focus, attend to certain parts of the input often for the purpose of selecting a relevant portion. Since attention mechanisms are also differentiable they are often jointly learned within a task. This perspective motivates our idea of a unification mechanism for learning variables across examples that utilises attention and is fully differentiable. We refer to this approach as \emph{soft unification} that can jointly learn which symbols can act as variables and how to assign values to them.

Hence, we propose an end-to-end differentiable neural network approach for learning and utilising the notion of lifting examples into invariants, which can then be used by the network to solve given tasks. The main contribution of this paper is a novel architecture capable of learning and using variables by lifting a given example through soft unification. As a first step, we present the empirical results of our approach in a controlled environment using four synthetic datasets and then with respect to a real-world dataset along with the analysis of the learned invariants that capture the underlying patterns present in the tasks. Our implementation using Chainer~\cite{chainer} is publicly available at
{\footnotesize \url{https://github.com/nuric/softuni}} with the accompanying data.

\section{Unification Networks}\label{sec:uni_nets}
Reasoning with variables involves first of all identifying what variables are in a given context as well as defining the process by which they are assigned values. The intuition is that when the varying components, i.e. variables, of an example are identified, the example can be lifted into an invariant that captures its structure but with variables replacing its varying components. Such an invariant can account for multiple other instances of the same structure. We present our approach in Algorithm~\ref{algo:uninet} and detail the steps below (note that the algorithm is more compressed than the steps described). Refer to Table~\ref{tab:notation_table} in Appendix~\ref{apx:model_details} for a summary of symbols and notations used in the paper.

\begin{step}[Pick Invariant Example]\label{step:pick}
We start from an example data point to generalise from. If we assume that within a task there is \emph{one} common pattern then any example should be an instance of that pattern. Therefore, we can randomly pick any example within a task from the dataset as our invariant example $G$. In this paper, $G$ is one data point consisting of a context, query and answer (see Table~\ref{tab:samples}).
\end{step}

\begin{step}[Lift Invariant Example]\label{step:lift}
In order for the invariant example to predict other examples correctly, certain symbols might need to \emph{vary} such as \vard{bernard} in Figure~\ref{fig:rule_babi_task16}. We capture this degree of \emph{variableness} with a function $\psi:\sS \rightarrow [0,1]$ for every symbol appearing in $G$ where $\sS$ is the set of all symbols. When $\psi$ is a learnable function, the model learns to identify the variables and convert the data point into an invariant, i.e. learning invariants. For a certain threshold $\psi(s) \geq t$, we visualise them in bold with a \var~prefix. An invariant is thus a pair $I \triangleq (G, \psi)$, the example data point and the variableness of each symbol. Which symbols emerge as variables depend on whether they need to be assigned new values, but how do we know which values to assign? This brings us to unification.
\end{step}

\begin{step}[Compute Unifying Features]\label{step:ufeats}
Suppose now we are given a new data point $K$ that we would like to unify with our invariant $I$ from the previous step. $K$ might start with a question ``what colour is lily'' and our invariant ``what colour is \vard{bernard}''. We would like to match bernard with lily. However, if we were to just use $d$-dimensional representations of symbols $\phi: \sS \rightarrow \R^d$, the representations of bernard and lily would need to be similar which might confound an upstream predictor network, e.g. when lily and bernard appear in the same story it would be difficult to distinguish them with similar representations. To resolve this issue, we learn \emph{unifying features} $\phi_U: \sS \rightarrow \R^d$ that intuitively capture some common meaning of two otherwise distinct symbols.
For example, $\phi(\text{bernhard})$ and $\phi(\text{lily})$ could represent specific people whereas $\phi_U(\text{bernhard}) \approx \phi_U(\text{lily})$ the notion of \emph{someone}; similarly, in Figure~\ref{fig:diag_softuni}, $\phi_U(7) \approx \phi_U(3)$ the notion of the head of a sequence. Notice how we use the original symbol bernhard in computing the representation of \vard{bernhard}; this is intended to capture the variable's bound \textit{meaning} following the idea of referants~\cite{senseandreference}.
%Although the person who invented computers and the person who shortened the Second World War might both refer to Alan Turing, they have different meanings. 
\end{step}

\setlength{\textfloatsep}{1em}
\begin{algorithm}[t]
\small
\DontPrintSemicolon
% ---------------------------
\SetKwFunction{ApplyMap}{map}
\SetKw{KwListIn}{in}
\SetKw{KwLet}{Let}
\SetKwComment{Comment}{$\triangleright$ }{}
\SetKwFunction{ExtractFeat}{extract}
% ---------------------------
\KwIn{Invariant $I$ consisting of example $G$ and variableness network $\psi$, example $K$, features network $\phi$, unifying features network $\phi_U$, upstream predictor network $f$}
\KwOut{Predicted label for example $K$}
\BlankLine
% ---------------------------
\Begin(\Comment*[h]{Unification Network}){
 \Return{$f \circ g(I, K)$}\Comment*[f]{Predictions using Soft Unification $g$}
}
\Begin(\Comment*[h]{Soft Unification function $g$}){
  \ForEach{symbol s \KwListIn G}{
    $\mA_{s,:} \leftarrow \phi(s)$\Comment*[r]{Features of $G$, $\mA \in \R^{|G|\times d}$}
    $\mB_{s,:} \leftarrow \phi_U(s)$\Comment*[r]{Unifying features of $G$, $\mB \in \R^{|G|\times d}$}
  }
  \ForEach{symbol s \KwListIn K}{
    $\mC_{s,:} \leftarrow \phi(s)$\Comment*[r]{Features of $K$, $\mC \in \R^{|K|\times d}$}
    $\mD_{s,:} \leftarrow \phi_U(s)$\Comment*[r]{Unifying features of $K$, $\mD \in \R^{|K|\times d}$}
  }
  \KwLet $\mP = \operatorname{softmax}(\mB\mD^T)$\Comment*[r]{Attention map over symbols, $\mP \in \R^{|G| \times |K|}$}\label{algoline:uninet_uni_att}
  \KwLet $\mE = \mP\mC$\Comment*[r]{Attended representations of $G$, $\mE \in \R^{|G|\times d}$}
  \ForEach{symbol s \KwListIn G}{
    $\mU_{s,:} \leftarrow \psi(s)\mE_{s,:} + (1-\psi(s))\mA_{s,:}$\Comment*[f]{Unified representation of $I$, $\mU \in \R^{|G| \times d}$}\label{algoline:uninet_inv_value}
  }
  \Return{$\mU$}
}
% ---------------------------
\caption{Unification Networks}
\label{algo:uninet}
\end{algorithm}

\begin{step}[Soft Unification]\label{step:unify}
Using learned unifying features, for every variable in $I$ we can now find a corresponding symbol in $K$. This process of unification, i.e. replacing variables with symbol values, is captured by the function $g$ that given an invariant and another example it updates the variables with appropriate values. To achieve this, we compute a soft attention for each symbol $s$ in the invariant using the unifying features $\phi_U(s)$ (line~\ref{algoline:uninet_uni_att} in Algorithm~\ref{algo:uninet}) and interpolate between its own and its variable value (line~\ref{algoline:uninet_inv_value} in Algorithm~\ref{algo:uninet}).
Since $g$ is a differentiable formulation and $\psi$, $\phi$ and $\phi_U$ can be learnable functions, we refer to this step as \emph{soft unification}. In Figure~\ref{fig:diag_softuni}, the variable \vard{7} is changed towards the symbol 3, having learnt that the unifiable feature is the head of the sequence.
\end{step}

\begin{step}[Predict]\label{step:predict}
So far we have constructed a unified data point of $I$ with the new example $K$ of which we would like to predict the answer. How do we predict? We use another, potentially upstream task specific, network $f$ that tries to predict the answer based on our unified input. Recall that our data points are triples of the form context, query and answer. In question answering, $f$ could be a memory network, or when working with grid like inputs, a CNN. By predicting on the output of our unification, we expect that $f \circ g(I,K) = f(K) = a$. If $f$ is differentiable, we can learn how to unify while solving the upstream task. We focus on $g$ and use standard networks for $f$ to understand which invariants are learned and the interaction of $f \circ g$ instead of the raw performance of $f$.
\end{step}

%\begin{figure}[h]
%\centering
%\includegraphics[width=0.9\linewidth]{diag/softuni.pdf}
%\caption{Graphical overview of soft unification. The output of the unification process $g$ is a weighted sum over the ground example using unifying features, equation~\ref{eq:inv_value}.}
%\label{fig:diag_softuni}
%\end{figure}

\section{Instances of Unification Networks}\label{sec:inst_uni_nets}
We present four architectures to model $f \circ g$ and demonstrate the flexibility of our approach towards different architectures and upstream tasks. Except in Unification RNN, the $d$-dimensional representation of symbols are learnable embeddings $\phi(s) = O[s]^T\mE$ with $\mE \in \R^{|\sS| \times d}$ randomly initialised by $\mathcal{N}(0,1)$ and $O[s]$ the one-hot encoding of the symbol. The variableness of symbols are learnable weights $\psi(s) = \sigma(w_s)$ where $\vw \in \R^{|\sS|}$ and $\sigma$ is the sigmoid function. We consider every symbol independently as a variable irrespective of its surrounding context and leave further contextualised formulations as future work. However, unifying features $\phi_U$ can be context sensitive to disambiguate same symbol variables appearing in different contexts. Full details the of models, including hyper-parameters, are available in Appendix~\ref{apx:model_details}.

\textbf{Unification MLP (UMLP)} ($f$: MLP, $g$: RNN) We start with a sequence of symbols as input, e.g. a sequence of digits \texttt{4234}. Unifying features $\phi_U$, from Step~\ref{step:ufeats}, are obtained using the hidden states of a bi-directional GRU~\cite{gru} processing the embedded sequences. In Step~\ref{step:predict}, the upstream MLP predicts the answer based on the flattened representation of the unified sequence.

\textbf{Unification CNN (UCNN)} ($f$: CNN, $g$: CNN) To adapt our approach for a grid of symbols, we use \emph{separate} convolutional neural networks with the same architecture to compute unifying features $\phi_U$ as well as to predict the correct answer through $f$ in Step~\ref{step:predict}. We mask out padding in Step~\ref{step:unify} to avoid assigning null values to variables. 

\textbf{Unification RNN (URNN)} ($f$: RNN, $g$: MLP) We start with a varying length sequence of words such as a movie review. We set ConceptNet word embeddings~\cite{numberbatch} as $\phi$ to compute $\phi_U(s) = \operatorname{MLP}(\phi(s))$ and $\psi(s) = \sigma(\operatorname{MLP}(\phi(s))$. Then, the final hidden state of $f$, an LSTM~\cite{lstm}, predicts the answer.

\begin{figure}[h]
\centering
\includegraphics[width=0.8\linewidth]{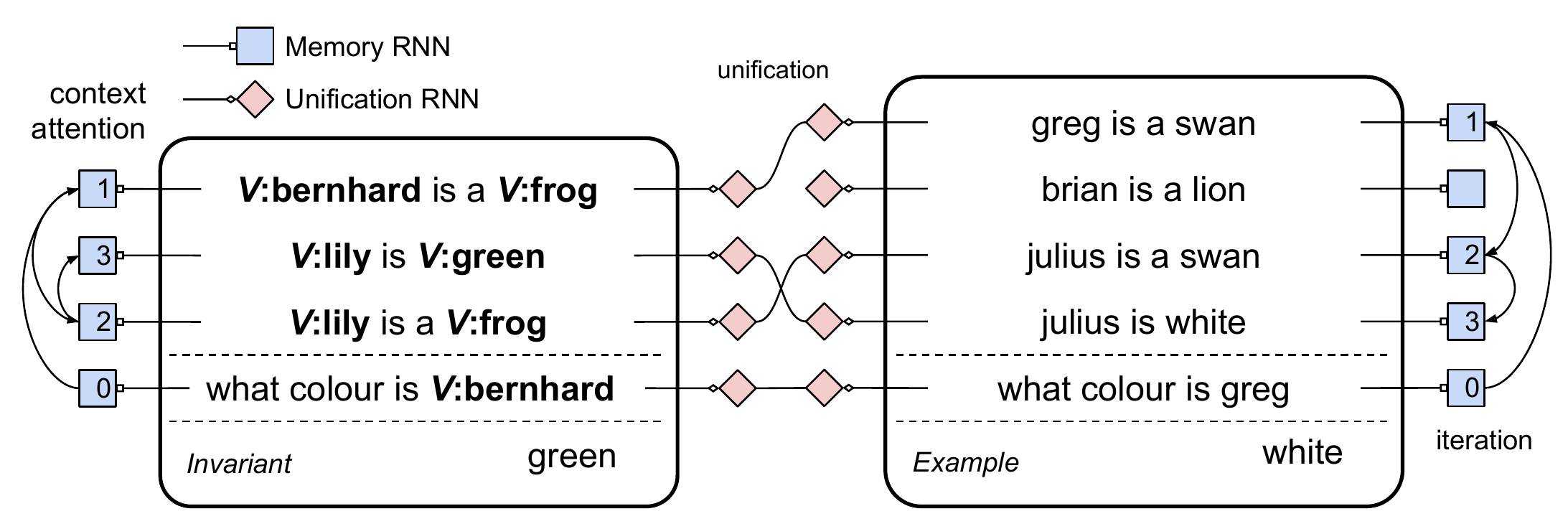}
\caption{Graphical overview of soft unification within a memory network (UMN). Each sentence is processed by two bi-directional RNNs for memory and unification. At each iteration the context attention selects which sentences to unify and the invariant produces the same answer as the example.}
\label{fig:diag_unimemnn}
\end{figure}
\textbf{Unification Memory Networks (UMN)} ($f$: MemNN, $g$: RNN) Soft unification does not need to happen prior to $f$ in a $f \circ g$ fashion but can also be incorporated at any intermediate stage multiple times. To demonstrate this ability, we unify the symbols at different memory locations at each iteration of a Memory Network~\cite{memnn}. We take a list of lists as input such as a tokenised story, Figure~\ref{fig:diag_unimemnn}. The memory network $f$ uses the final hidden state of a bi-directional GRU (blue squares in Figure~\ref{fig:diag_unimemnn}) as the sentence representations to compute a context attention, i.e. select the next context sentence starting with the query. With the sentences attended to, we can unify the words of the sentences at each iteration following Steps~\ref{step:lift} to \ref{step:unify}. We use another bi-directional GRU (pink diamonds in Figure~\ref{fig:diag_unimemnn}) for unifying features $\phi_U$. Following line~\ref{algoline:uninet_inv_value} in Algorithm~\ref{algo:uninet}, the new unified representation of the memory slot (the sentence) is used by $f$ to perform the next iteration. The prediction is then based on the final hidden state of the invariant example. This setup, however, requires pre-training $f$ such that the context attentions match the correct pairs of sentences to unify which limits the performance of the combined network by how well $f$ performs.

Although we assume a single pattern in Step~\ref{step:pick}, a task might contain slightly different examples such as ``Where is X?'' and ``Why did X go to Y?''. To let the models potentially learn and benefit from different invariants, we can pick multiple examples to generalise from and aggregate the predictions from each invariant. One simple approach is to sum the predictions of the invariants $\sum_{I \in \sI} f \circ g(I,K)$ used in UMLP, UCNN and URNN where $\sI$ is the set of invariants. For UMN, at each iteration we weigh the hidden states from each invariant using a bilinear attention $\eta = \text{softmax}(\vh^0_I \mW \vh^{0^T}_K)$ where $\vh^0_I$ and $\vh^0_K$ are the representations of the query (at iteration 0).

\begin{table*}[h]
\small
\centering
\caption{Sample context, query and answer triples and their training sizes \emph{per task}. For the distribution of generated number of examples per task on Sequence and Grid data refer to Appendix~\ref{apx:dataset}.}
\label{tab:samples}
\begin{tabular}{@{}ccccc@{}}
\toprule
Dataset & Context & Query & Answer & Training Size\\
\midrule
Sequence & 8384 & duplicate & 8 & $\le$ 1k, $\le$ 50\\
Grid &
\setlength{\tabcolsep}{1mm}
\renewcommand{\arraystretch}{0.5}
\begin{tabular}{@{}ccc@{}}
 0 & 0 & 3\\
 0 & 1 & 6\\
 8 & 5 & 7
\end{tabular} & corner & 7 & $\le$ 1k, $\le$ 50\\
bAbI &
\begin{tabular}{@{}l@{}}
Mary went to the kitchen.\\
Sandra journeyed to the garden.
\end{tabular} & Where is Mary? & kitchen & 1k, 50\\
Logic &
\begin{tabular}{@{}l@{}}
p(X) $\leftarrow$ q(X).\\
q(a).
\end{tabular} & p(a). & True & 2k, 100\\
Sentiment A. & easily one of the best films & Sentiment & Positive & 1k, 50\\
\bottomrule
\end{tabular}
\end{table*}

\section{Datasets}\label{sec:datasets}

We use five datasets consisting of context, query and an answer $(C, Q, a)$ (see Table~\ref{tab:samples} and Appendix~\ref{apx:dataset} for further details) with varying input structures: fixed or varying length sequences, grids and nested sequences (e.g. stories). In each case we use an appropriate model: UMLP for fixed length sequences, UCNN for grid, URNN for varying length sequences and UMN for iterative reasoning.
%We mainly use synthetic datasets of which the data generating distributions are known to evaluate not only the quantitative performance but also the quality of the invariants learned by our approach.
%Therefore, to provide a qualitative analysis of the proposed approach and to facilitate reproducible results, we refrain from using \emph{real-world} datasets and opt for a controlled environment with synthetic datasets for experimentation.

\begin{wrapfigure}{r}{0.50\linewidth}
\centering
\includegraphics[width=0.9\linewidth]{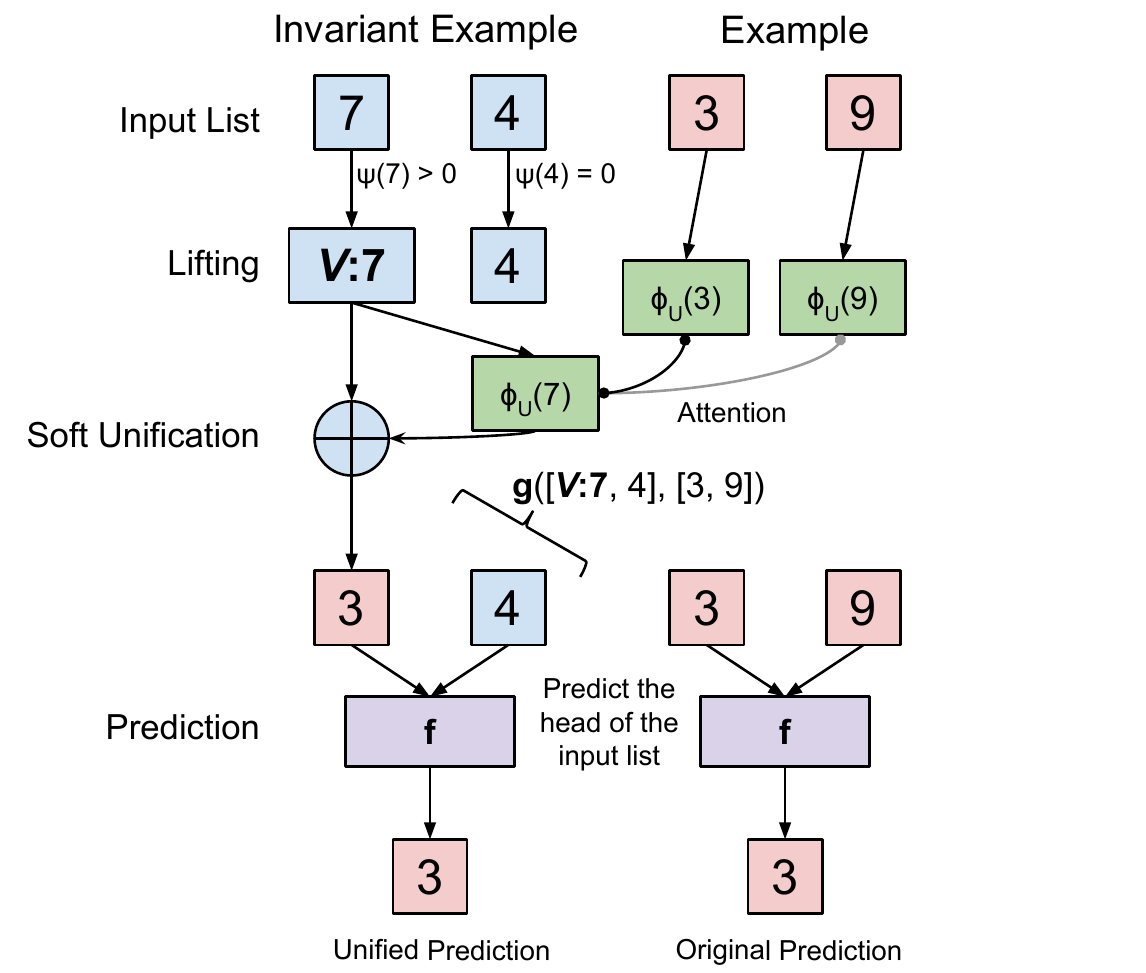}
\caption{Graphical overview of predicting the head of a sequence using an invariant and soft unification where $g$ outputs the new sequence 3 4.}
\label{fig:diag_softuni}
\end{wrapfigure}

\textbf{Fixed Length Sequences} We generate sequences of length $l=4$ from 8 unique symbols represented as digits to predict (i) a constant, (ii) the head of the sequence, (iii) the tail and (iv) the duplicate symbol. We randomly generate 1000 triples and then only take the unique ones to ensure the test split contains unseen examples. The training is then performed over a 5-fold cross-validation. Figure~\ref{fig:diag_softuni} demonstrates how the invariant `\vard{7} 4' can predict the head of another example sequence `3 9'.

\textbf{Grid} To spatially organise symbols, we generate a grid of size $3 \times 3$ from 8 unique symbols. The grids contain one of (i) $2 \times 2$ box of identical symbol, (ii) a vertical, diagonal or horizontal sequence of length 3, (iii) a cross or a plus shape and (iv) a triangle. In each task we predict (i) the identical symbol, (ii) the head of the sequence, (iii) the centre of the cross or plus and (iv) the corner of the triangle respectively. We generate 1000 triples discarding any duplicates.

\textbf{bAbI} The bAbI dataset consists of 20 synthetically generated natural language reasoning tasks (refer to~\cite{babi} for task details). We take the 1k English set and use 0.1 of the training set as validation. Each token is lower cased and considered a unique symbol. Following previous works~\cite{qrn, memn2n}, we take multiple word answers also to be a unique symbol. To initially form the repository of invariants, we use the bag-of-words representation of the questions and find the most dissimilar ones based on their cosine similarity as a heuristic to obtain varied examples.

\textbf{Logical Reasoning} To distinguish our notion of a variable from that used in logic-based formalisms, we generate logical reasoning tasks in the form of logic programs using the procedure from~\cite{deeplogic}. The tasks involve learning $f(C,Q) = \text{True}$ if and only if $C \vdash Q$ over 12 classes of logic programs exhibiting varying paradigms of logical reasoning including negation by failure~\cite{nbf}. We generate 1k and 50 logic programs per task for training with 0.1 as validation and another 1k for testing. Each logic program has one positive and one negative prediction giving a total of 2k and 100 data points respectively. We use one random character from the English alphabet for predicates \emph{and} constants, e.g. $p(p,p)$ and an upper case character for logical variables, e.g. $p(\text{\texttt{X}},\text{\texttt{Y}})$. Further configurations such as restricting the arity of predicates to 1 are presented in Table~\ref{tab:deeplogic_results} Appendix~\ref{apx:further_results}.

\textbf{Sentiment Analysis} To evaluate on a noisy real-world dataset, we take the sentiment analysis task from~\cite{sochersentiment} and prune sentences to a maximum length of 20 words. We threshold the scores $\le 0.1$ and $\ge 0.9$ for negative and positive labels respectively to ensure unification cannot yield a neutral score, i.e. the model is forced to learn either a positive or a negative label. We then take 1000 or 50 training examples \emph{per label} and use the remaining $\approx 676$ data points as unseen test examples.

\section{Experiments}
We probe three aspects of soft unification: the impact of unification on performance over unseen data, the effect of multiple invariants and data efficiency. To that end, we train UMLP, UCNN and URNN with and without unification and UMN with pre-training using 1 or 3 invariants over either the entire training set or only 50 examples.
Every model is trained via back-propagation using Adam~\cite{adam} with learning rate 0.001 on an Intel Core i7-6700 CPU using the following objective function:
\begin{equation}\label{eq:objective_function}
J = \overbrace{\lambda_K \nll(f)}^\text{Original output} + \lambda_I [ \ \overbrace{\nll(f \circ g)}^\text{Unification output} + \overbrace{\tau \sum_{s \in \sS} \psi(s) }^\text{Sparsity} \ ]
\end{equation}
where $\nll$ is the negative log-likelihood with sparsity regularisation over $\psi$ at $\tau = 0.1$ to discourage the models from utilising spurious number of variables.
We add the sparsity constraint over the variableness of symbols $\psi(s)$ to avoid the trivial solution in which every symbol is a variable and $G$ is completely replaced by $K$ still allowing $f$ to predict correctly. Hence, we would like the \emph{minimal} transformation of $G$ towards $K$ to expose the common underlying pattern.
For UMLP and UCNN, we set $\lambda_K = 0,\ \lambda_I = 1$ for training just the unified output and the converse for the non-unifying versions. For URNN, we set $\lambda_K = \lambda_I = 1$ to train the unified output and set $\lambda_I=0$ for non-unifying version. To pre-train the UMN, we start with $\lambda_K = 1,\ \lambda_I = 0$ for 40 epochs then set $\lambda_I = 1$ to jointly train the unified output. For UMN, we also add the mean squared error between hidden states of $I$ and $K$ at each iteration (see Appendix~\ref{apx:training_details}). In the strongly supervised cases, the negative log-likelihood of the context attentions (which sentences are selected at each iteration) are also added. Further training details including sample training curves are available in Appendix~\ref{apx:training_details}.

\begin{figure*}[t]
\centering
\includegraphics[width=1.0\textwidth]{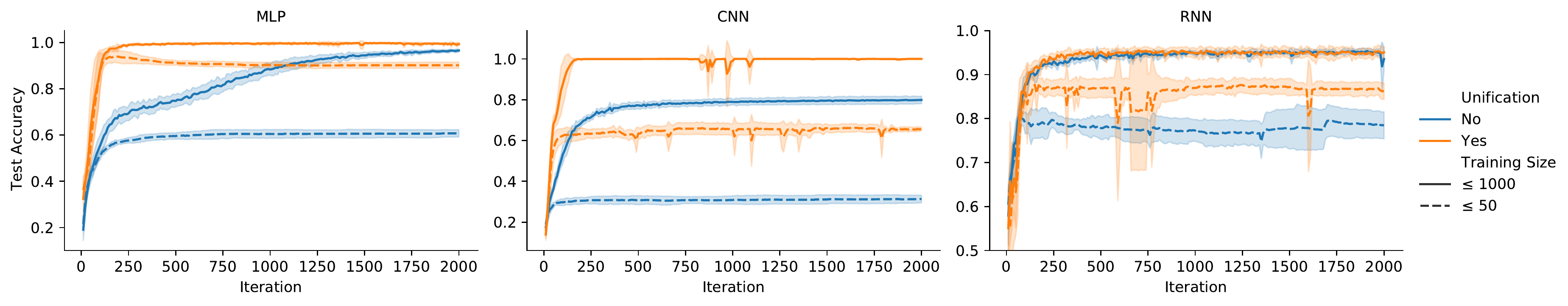}
\caption{Test accuracy over iterations for UMLP, UCNN and URNN models with 1 invariant versus no unification. Soft unification aids with data efficiency of models against their plain counterparts.}
\label{fig:umlp_ucnn_acc}
\end{figure*}

Figure~\ref{fig:umlp_ucnn_acc} portrays how soft unification generalises better to unseen examples in test sets over plain models. Despite $f \circ g$ having more trainable parameters than $f$ alone, this data efficiency is visible across all models when trained with only $\le 50$ examples \emph{per task}. We believe soft unification architecturally biases the models towards learning unifying features that are common across examples, therefore, potentially also common to unseen examples. The data efficient nature is more emphasised with UMLP and UCNN on synthetic datasets in which there are unambiguous patterns in the tasks and they achieve higher accuracies in as few as 250 iterations (batch updates) against their plain counterparts. In the real-world dataset of sentiment analysis, we observe a less steady training curve for URNN and performance as good as if not better than its plain version. The fluctuations in accuracy around iterations 750 to 1000 in UCNN and iteration 700 in URNN are caused by penalising $\psi$ which forces the model to adjust the invariant to use less variables half way through training. Results with multiple invariants are identical and the models learn to ignore the extra invariants (Figure~\ref{fig:umlp_ucnn_invs} Appendix~\ref{apx:further_results}) due to the regularisation applied on $\psi$ zeroing out unnecessary invariants. Training with different learning rates overall paint a similar picture (Figure~\ref{fig:umlp_ucnn_learning_rates} Appendix~\ref{apx:training_details}).

\begin{table*}[h]
\footnotesize
\centering
\caption{Aggregate error rates (\%) on bAbI 1k for UMN and baselines N2N~\cite{memn2n}, GN2N~\cite{gn2n}, EntNet~\cite{entnet}, QRN~\cite{qrn} and MemNN~\cite{memnn} respectively. Full comparison is available in Appendix~\ref{apx:further_results}.}
\label{tab:babi_agg_results}
\begin{tabular}{@{}rccccc|ccccc@{}}
\toprule
Training Size   & \multicolumn{4}{c|}{1k} & 50 & \multicolumn{5}{c}{1k}\\
Supervision & \multicolumn{2}{c|}{Weak} & \multicolumn{3}{c|}{Strong} & \multicolumn{4}{c|}{Weak} & Strong\\
\# Invs / Model & 1 & 3 & 1 & 3 & 3 & N2N & GN2N & EntNet & QRN & MemNN\\
\midrule
Mean & 18.8 & 19.0 & \textbf{5.1} & 6.6 & 28.7 & 13.9 & 12.7 & 29.6 & 11.3 & 6.7\\
\# $>5\%$ & 10 & 9 & 3 & 3 & 17 & 11 & 10 & 15 & 5 & 4\\
\bottomrule
\end{tabular}
\end{table*}

\begin{table*}[b]
\footnotesize
\centering
\caption{Aggregate task error rates (\%) on the logical reasoning dataset for UMN and baseline IMA~\cite{deeplogic}. Individual task results are available in Appendix~\ref{apx:further_results}.}
\label{tab:deeplogic_agg_results}
\begin{tabular}{@{}rccccc|cc@{}}
\toprule
Model & \multicolumn{5}{c|}{UMN} & \multicolumn{2}{c}{IMA}\\
Training Size & \multicolumn{4}{c|}{2k} & 100 & \multicolumn{2}{c}{2k}\\
Supervision  & \multicolumn{2}{c|}{Weak} & \multicolumn{3}{c|}{Strong} & Weak & Strong\\
\# Invs & 1 & 3 & 1 & 3 & 3 & \multicolumn{2}{c}{-}\\
\midrule
Mean & 37.7 & 37.6 & \textbf{27.4} & 29.0 & 47.1 & 38.8 & 31.5\\
\# $>5\%$ & 10 & 10 & 10 & 11 & 12 & 11 & 11\\
\bottomrule
\end{tabular}
\end{table*}

For iterative reasoning tasks, Tables~\ref{tab:babi_agg_results} and \ref{tab:deeplogic_agg_results} aggregate the results for our approach against comparable baseline memory networks which are selected based on whether they are built on Memory Networks (MemNN)~\cite{memnn} and predict by iteratively updating a hidden state. For example, End-to-End Memory Networks (N2N)~\cite{memn2n} and Iterative Memory Attention (IMA)~\cite{deeplogic} networks update a hidden state vector after each iteration by attending to a single context sentence similar to our architecture. We observe that strong supervision and more data per task yield lower error rates which is consistent with previous work reflecting how $f \circ g$ can be bounded by the efficacy of $f$ modelled as a memory network. In a weak supervision setting, i.e. when sentence selection is not supervised, our model attempts to unify arbitrary sentences often failing to follow the iterative reasoning chain. As a result, only in the supervised case we observe a minor improvement over MemNN by 1.6 in Table~\ref{tab:babi_agg_results} and over IMA by 4.1 in Table~\ref{tab:deeplogic_agg_results}. Without reducing the performance of $f$, our approach is still able to learn invariants as shown in Figure~\ref{fig:invariants}. This dependency on $f$ also limits the ability of $f \circ g$ to learn from 50 and 100 examples per task failing 17/20 of bAbI and 12/12 of logical reasoning tasks respectively. The increase in error rate with 3 invariants in Table~\ref{tab:deeplogic_agg_results}, we speculate, stems from having more parameters and more pathways, rendering training more difficult and slower.

\begin{table}[h]
\small
\centering
\caption{Number of \emph{exact} matches of the learnt invariants with expected ones in synthetic datasets where the invariant is known. If the model does not get an exact match, it might use more or fewer variables than expected while still solving the task. For sequences and grid datasets, there are 5 folds each with 4 tasks giving 20 and the logic dataset has 12 tasks with 3 runs giving 36 invariants.}
\label{tab:correctness}
\begin{tabular}{@{}rcccccc@{}}
\toprule
Model &  \multicolumn{2}{c}{UMLP} & \multicolumn{2}{c}{UCNN} & \multicolumn{2}{c}{UMN}\\
Dataset & \multicolumn{2}{c}{Sequences} & \multicolumn{2}{c}{Grid} & \multicolumn{2}{c}{Logic}\\
Train Size & $\leq1k$ & $\leq50$ & $\leq1k$ & $\leq50$ & \multicolumn{2}{c}{2k}\\
Supervision & \multicolumn{4}{c}{-} & Weak & Strong\\
\midrule
Correct / Total & 18/20 & 18/20 & 13/20 & 14/20 & 7/36 & 23/36\\
Accuracy (\%) & 90.0 & 90.0 & 65.0 & 70.0 & 19.4 & 63.9\\
\bottomrule
\end{tabular}
\end{table}

For synthetic sequences, grid and logic datasets in which we know exactly what the invariants \emph{can be}, Table~\ref{tab:correctness} shows how often our approach captures \emph{exactly} the expected invariant. We threshold $\psi$ as explained in Section~\ref{sec:analysis} and check for an exact match; for example for predicting the head of a sequence in UMLP, we compare the learnt invariant against the pattern ``\var \_ \_ \_''. Although with increasing dataset complexity the accuracy drops, it is important to note that just because the model does not capture the exact invariant it may still solve the task. In these cases, it may use extra or more interestingly fewer variables as further discussed in Section~\ref{sec:analysis}.

\section{Analysis}\label{sec:analysis}
\begin{wrapfigure}{r}{0.4\linewidth}
\centering
\includegraphics[width=1.0\linewidth]{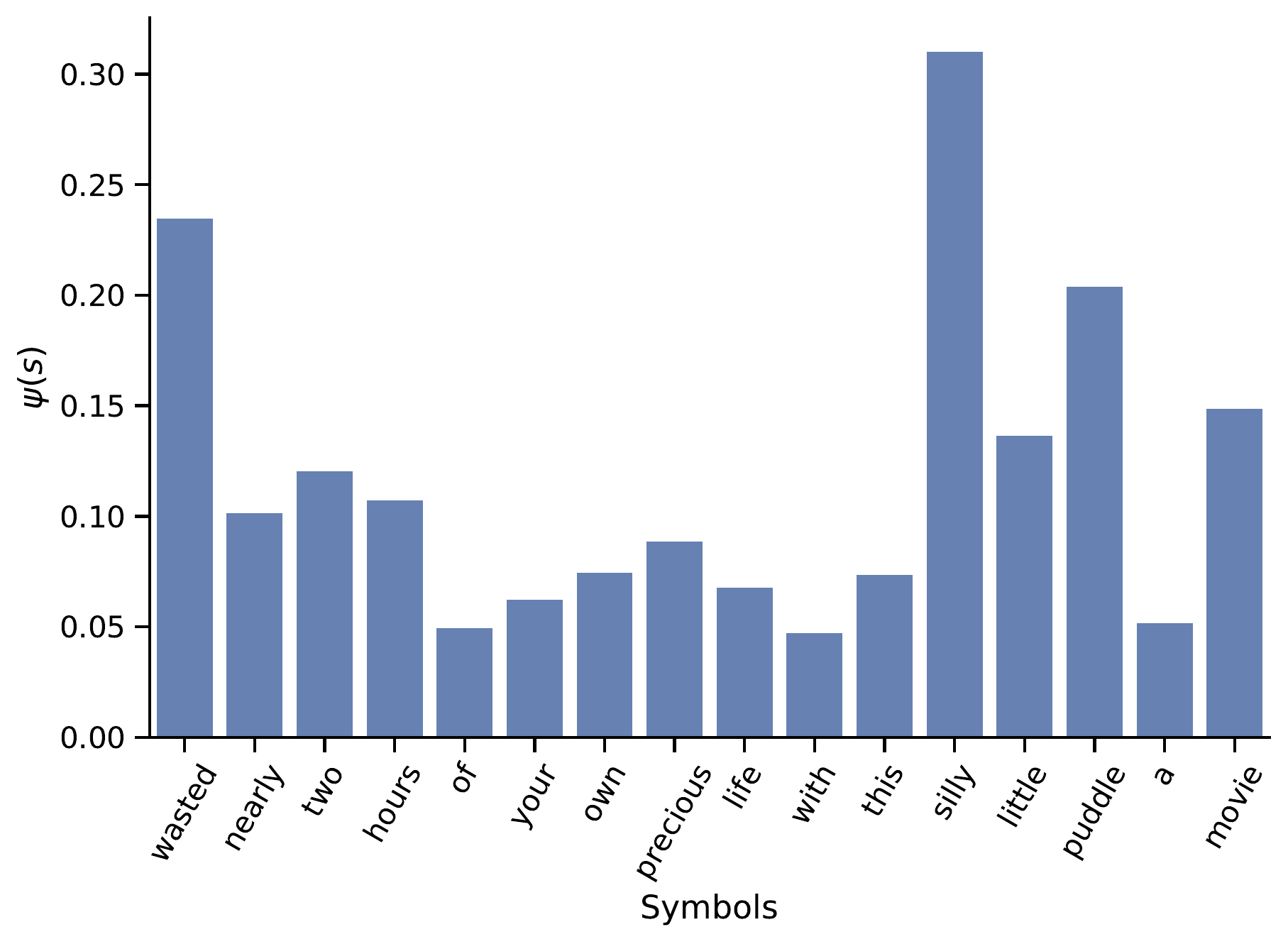}
\caption{The variableness $\psi(s)$ of an invariant for sentiment analysis with words such as `silly' emerging as variables.}
\label{fig:sentiment_inv}
\end{wrapfigure}
Figure~\ref{fig:sentiment_inv} shows an invariant for sentiment analysis in which words such as silly that contribute more to the sentiment have a higher $\psi$. Intuitively, if one replaces `silly' with the adjective `great', the sentiment will change. The replacement, however, is not a hard value assignment but an interpolation (line~\ref{algoline:uninet_inv_value} Algorithm~\ref{algo:uninet}) which may produce a new intermediate representation from $G$ towards $K$ different enough to allow $f$ to predict correctly. Since we penalise the magnitude of $\psi$ in equation~\ref{eq:objective_function}, we expect these values to be as low as possible. For synthetic datasets, we apply a threshold $\psi(s) > t$ to extract the learned invariants and set $t$ to be the mean of the variable symbols as a heuristic except for bAbI where we use $t=0.1$. The magnitude of $t$ depends on the amount of regularisation $\tau$, equation~\ref{eq:objective_function}, number of iterations and batch size. Sample invariants shown in Figure~\ref{fig:invariants} describe the patterns present in the tasks with parts that contribute towards the final answer becoming variables. Extra symbols such as `is' or `travelled' do not emerge as variables, as shown in Figure~\ref{fig:rule_babi_2}; we attribute this behaviour to the fact that changing the token `travelled' to `went' does not influence the prediction but changing the action, the value of \vard{left} to `picked' does. However, based on random initialisation, our approach can convert an arbitrary symbol into a variable and let $f$ compensate for the unifications it produces. For example, the invariant ``\vard{8} 5 2 2'' could predict the tail of another example by unifying the head with the tail using $\phi_U$ of those symbols in Step~\ref{step:ufeats}. Further examples are shown in Appendix~\ref{apx:further_results}. Pre-training $f$ as done in UMN seems to produce more robust and consistent invariants since, we speculate, a pre-trained $f$ encourages more $g(I,K) \approx K$.

\begin{figure*}[t]
\footnotesize
\centering
\begin{subfigure}[t]{0.48\textwidth}
\centering
\renewcommand{\arraystretch}{1.2}
\begin{tabular}{l}
  \vard{john} travelled to the \vard{office}\\
  \vard{john} \vard{left} the \vard{football}\\
  \hline
  where is the \vard{football}\\
  \hline
  office
\end{tabular}
\caption{bAbI task 2, two supporting facts. The model also learns \vard{left} since people can also drop or pick up objects potentially affecting the answer.}
\label{fig:rule_babi_2}
\end{subfigure}
~
\begin{subfigure}[t]{0.49\textwidth}
\centering
\renewcommand{\arraystretch}{1.2}
\begin{tabular}{l}
  this \vard{morning} \vard{bill} went to the \vard{school}\\
  yesterday \vard{bill} journeyed to the \vard{park}\\
  \hline
  where was \vard{bill} before the \vard{school}\\
  \hline
  park
\end{tabular}
\caption{bAbI task 14, time reasoning. \vard{bill} and \vard{school} are recognised as variables alongside \vard{morning} capturing \emph{when} someone went.}
\label{fig:rule_babi_14}
\end{subfigure}\\[1em]
\begin{subfigure}[t]{0.31\textwidth}
\centering
\setlength{\tabcolsep}{0.2em}
\begin{tabular}{rcr}
%umlp_result/umlp_l4_s8_i1_e16_t50_f3.out
5 8 6 4 & const & 2\\
\vard{8} 3 3 1 & head & 8\\
8 3 1 \vard{5} & tail & 5\\
\vard{1} 4 3 \vard{1} & dup & 1
\end{tabular}
\caption{Successful invariants learned with UMLP using 50 training examples only shown as $(C,Q,a)$.}
\label{fig:rule_seq_tasks}
\end{subfigure}
~
\begin{subfigure}[t]{0.32\textwidth}
\centering
\begin{tabular}{ccc}
%ucnn_result/ucnn_s8_i1_e32_t0_f1.out
\setlength{\tabcolsep}{0.2em}
\begin{tabular}{@{}rrr@{}}
 0 & \var & \var\\
 0 & \var & \var\\
 0 & 0 & 0
\end{tabular} &
\setlength{\tabcolsep}{0.2em}
\begin{tabular}{@{}rrr@{}}
 0 & 1 & 0\\
 6 & \var & 8\\
 0 & 7 & 0
\end{tabular} &
\setlength{\tabcolsep}{0.2em}
\begin{tabular}{@{}rrr@{}}
 0 & 0 & 1\\
 0 & 5 & 4\\
 7 & 8 & \var
\end{tabular}\\
box & centre & corner
\end{tabular}
\caption{Successful invariants learned with UCNN. Variable default symbols are omitted for clarity.}
\label{fig:rule_grid_tasks}
\end{subfigure}
~
\begin{subfigure}[t]{0.31\textwidth}
\centering
\begin{tabular}{l}
  \vard{i} ( \texttt{T} ) $\leftarrow$ \vard{l} ( \texttt{T} ),\\
  \vard{l} ( \texttt{U} ) $\leftarrow$ \vard{x} ( \texttt{U} ),\\
  \vard{x} ( \texttt{K} ) $\leftarrow$ \vard{n} ( \texttt{K} ),\\
  \vard{n} ( \vard{o} ) $\vdash$ \vard{i} ( \vard{o} )
\end{tabular}
\caption{Logical reasoning task 5 with arity 1. The model captures how \vard{n} could entail \vard{i} in a chain.}
\label{fig:rule_deeplogic_5}
\end{subfigure}
\caption{Invariants learned across the four datasets using the three architectures. For iterative reasoning datasets, bAbI and logical reasoning, they are taken from strongly supervised UMN.}
%\label{fig:qaxx_r1_s1_t0_s0}
%\label{fig:dlxx_arx_10k_r1_s1_t0_s0}
\label{fig:invariants}
\end{figure*}

\textbf{Interpretability versus Ability} A desired property of interpretable models is transparency~\cite{interpretability}. A novel outcome of the learned invariants in our approach is that they provide an approximation of the underlying general principle that may be present in the data. Figure~\ref{fig:rule_deeplogic_5} captures the structure of multi-hop reasoning in which a predicate \vard{n} can entail another \vard{i} with a matching constant \vard{o} if there is a chain of rules that connect the two predicates. However, certain aspects regarding the ability of the model such as how it performs temporal reasoning, are still hidden inside $f$. In Figure~\ref{fig:rule_babi_14}, although we observe \vard{morning} as a variable, the overall learned invariant captures nothing about how changing the value of \vard{morning} alters the behaviour of $f$, i.e. how $f$ uses the interpolated representations produced by $g$. The upstream model $f$ might look \emph{before} or \emph{after} a certain time point \vard{bill} went somewhere depending what \vard{morning} binds to. Without the regularising term on $\psi(s)$, we initially noticed the models using, what one might call extra, symbols as variables and binding them to the same value occasionally producing unifications such as ``bathroom bathroom to the bathroom'' and still $f$ predicting, unsurprisingly, the correct answer as bathroom. Hence, regularising $\psi$ with the correct amount $\tau$ in equation~\ref{eq:objective_function} to reduce the capacity of unification seems critical in extracting not just any invariant but one that represents the common structure.

\begin{figure}[h]
\centering
\begin{subfigure}[t]{0.30\textwidth}
\centering
\includegraphics[width=1.00\textwidth,height=8em,keepaspectratio]{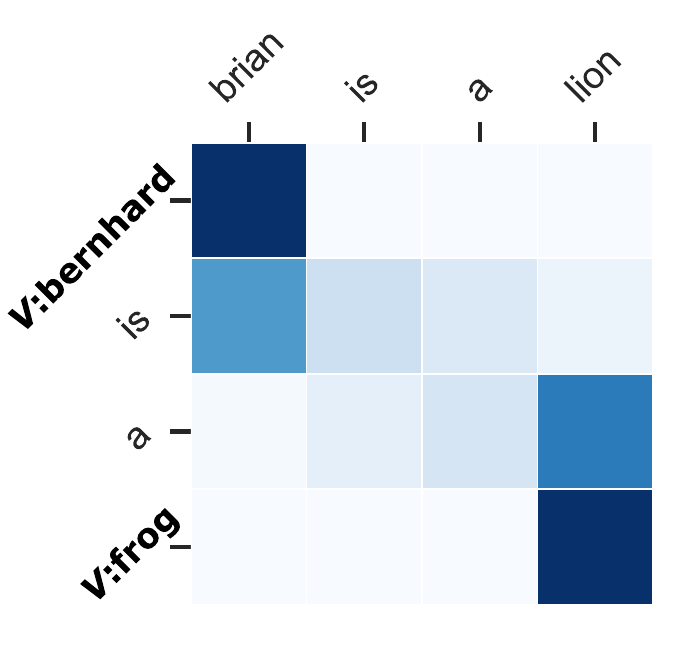}
\caption{bAbI task 16. A one-to-one mapping is created between variables \vard{bernhard} with brian and \vard{frog} with lion.}
\label{fig:uni_babi_16}
\end{subfigure}
~
\begin{subfigure}[t]{0.30\textwidth}
\centering
\raisebox{-1em}{\includegraphics[width=1.00\textwidth,height=10em,keepaspectratio]{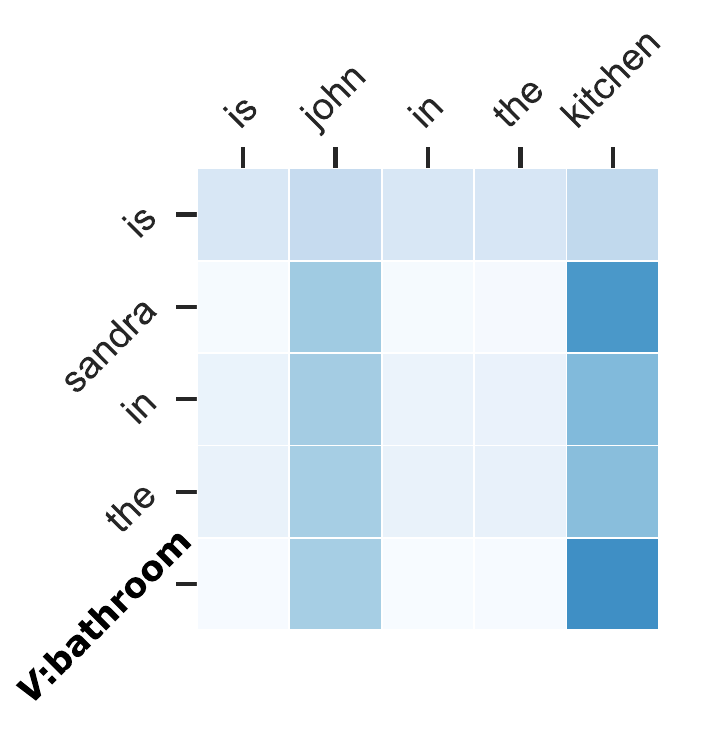}}
\vspace{-1em}
\caption{bAbI task 6. \vard{bathroom} is recognised as the only variable creating a one-to-many binding to capture the same information.}
\label{fig:uni_babi_06}
\end{subfigure}
~
\begin{subfigure}[t]{0.30\textwidth}
\centering
\includegraphics[width=0.63\textwidth,height=8em,keepaspectratio]{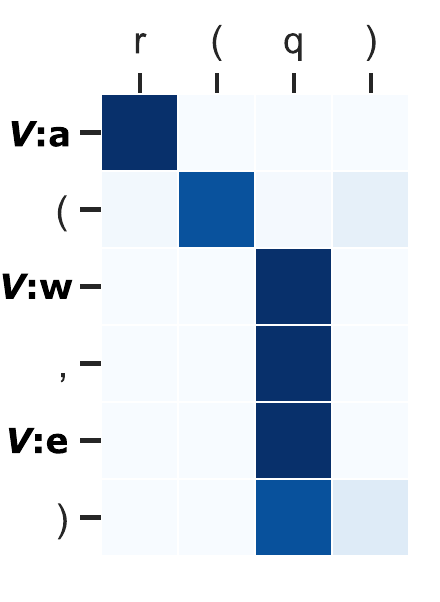}
\caption{Logical reasoning task 1. An arity 2 predicate is forced to bind with arity 1 creating a many-to-one binding.}
\label{fig:uni_dl_01_ar2_to_ar1}
\end{subfigure}
\caption{Variable bindings from line~\ref{algoline:uninet_uni_att} in Algorithm~\ref{algo:uninet}. Darker cells indicate higher attention values.}
\label{fig:uni}
\end{figure}

Attention maps from line~\ref{algoline:uninet_uni_att} in Algorithm~\ref{algo:uninet} reveal three main patterns: one-to-one, one-to-many or many-to-one bindings as shown in Figure~\ref{fig:uni} with more in Appendix~\ref{apx:further_results}. Figure~\ref{fig:uni_babi_16} captures what one might expect unification to look like where variables unify with their corresponding counterparts, e.g. \vard{bernhard} with brian and \vard{frog} with lion. However, occasionally the model can optimise to use less variables and \emph{squeeze} the required information into a single variable, for example by binding \vard{bathroom} to john and kitchen as shown in Figure~\ref{fig:uni_babi_06}. We believe this occurs due to the sparsity constraint on $\psi(s)$ encouraging the model to be as conservative as possible. Since the upstream network $f$ is also trained, it has the capacity to compensate for condensed or malformed unified representations; a possible option could be to freeze the upstream network while learning to unify. Finally, if there are more variables than needed as in Figure~\ref{fig:uni_dl_01_ar2_to_ar1}, we observe a many-to-one binding with \vard{w} and \vard{e} mapping to the same constant $q$. This behaviour begs the question how does the model differentiate between $p(q)$ and $p(q,q)$. We speculate the model uses the magnitude of $\psi(w) = 0.037$ and $\psi(e) = 0.042$ to encode the difference despite both variables unifying with the same constant.

\section{Related Work}
Learning an underlying general principle in the form of an invariant is often the means for arguing for generalisation in neural networks. For example, Neural Turing Machines~\cite{neuralturing} are tested on previously unseen sequences to support the view that the model might have captured the underlying pattern or algorithm. In fact, \cite{memnn} claim ``MemNNs can discover simple linguistic patterns based on verbal forms such as (X, dropped, Y), (X,
took, Y) or (X, journeyed to, Y) and can successfully generalise the meaning of their instantiations.'' However, this claim is based on the output of $f$ and unfortunately it is unknown whether the model has truly learned such a representation or indeed is utilising it. Our approach sheds light on this ambiguity and presents these linguistic patterns explicitly as invariants, ensuring their utility through $g$ without solely analysing the output of $f$ on previously unseen symbols. Although we associate these invariants with our existing understanding of the task to perhaps mistakenly anthropomorphise the machine, for example by thinking it has learned \vard{mary} as \emph{someone}, it is important to acknowledge that these are just symbolic patterns. They do not make our model, in particular $f$, more interpretable in terms of how these invariants are used or what they mean to the model. In these cases, our interpretations may not necessarily correspond to any understanding of the machine, relating to the Chinese room argument~\cite{chineseroom}.

Learning invariants by lifting ground examples is related to least common generalisation~\cite{leastcommongeneraliser} by which inductive inference is performed on facts~\cite{lcgfromfacts} such as generalising \textit{went(mary,kitchen)} and \textit{went(john,garden)} to \textit{went(X,Y)}. Unlike in a predicate logic setting, our approach allows for soft alignment and therefore generalisation between varying length sequences. Existing neuro-symbolic systems~\cite{neurosymbolic} focus on inducing rules that adhere to \emph{given} logical semantics of what variables and rules are. For example, $\delta ILP$~\cite{evans2018learning} constructs a network by rigidly following the given semantics of first-order logic. Similarly, Lifted Relational Neural Networks~\cite{liftedneuralnetworks} ground first-order logic rules into a neural network while Neural Theorem Provers~\cite{timntp} build neural networks using backward-chaining~\cite{russell2016artificial} on a given background knowledge base with templates.
%This architectural approach for combining logical variables is also observed with TensorLog~\cite{tensorlog} and Logic Tensor Networks~\cite{logictensor}.
%while grounding logical rules can also be used as regularisation~\cite{harnessinglogicregularisation}.
However, the notion of a variable is pre-defined rather than learned with a focus on presenting a practical approach to solving certain problems, whereas our motivation stems from a cognitive perspective.

At first it may seem the learned invariants, Section~\ref{sec:analysis}, make the model more interpretable; however, this transparency is not of the model $f$ but of the data. The invariant captures patterns that potentially approximates the data generating distribution but we still do not know \emph{how} the model $f$ uses them upstream. Thus, from the perspective of explainable artificial intelligence (XAI)~\cite{xaisurvey}, learning invariants or interpreting them does not constitute an explanation of the reasoning model $f$ even though ``if \emph{someone} goes \emph{somewhere} then they are there'' might look like one. Instead, it can be perceived as causal attribution~\cite{xaisocialsci} in which someone being somewhere is attributed to them going there. This perspective also relates to gradient based model explanation methods such as Layer-Wise Relevance Propagation~\cite{layerwiserelevance} and Grad-CAM~\cite{gradcam,gradcamplus}. Consequently, a possible view on $\psi$, Section~\ref{sec:uni_nets}, is a gradient based usefulness measure such that a symbol utilised upstream by $f$ to determine the answer becomes a variable similar to how a group of pixels in an image contribute more to its classification. However, gradient based saliency methods have shown to be unreliable if based solely on visual assessment~\cite{sanitychecksforsaliencymaps}.

Finally, one can argue that our model maintains a form of counterfactual thinking~\cite{counterfactualthinking} in which soft unification $g$ creates counterfactuals on the invariant example to alter the output of $f$ towards the desired answer, Step~\ref{step:predict}. The question \emph{where Mary would have been if Mary had gone to the garden instead of the kitchen} is the process by which an invariant is learned through multiple examples during training. This view relates to methods of causal inference~\cite{pearlcausalinference,statisticscausalinference} in which counterfactuals are vital as demonstrated in structured models~\cite{pearlcausationcounterfactual}.

\section{Conclusion}
We presented a new approach for learning variables and lifting examples into invariants through the usage of soft unification. Application of our approach to five datasets demonstrates that Unification Networks perform comparatively if not better to existing architectures without soft unification while having the benefit of lifting examples into invariants that capture underlying patterns present in the tasks.
Since our approach is end-to-end differentiable, we plan to apply this technique to multi-modal tasks in order to yield multi-modal invariants for example in visual question answering.

\subsubsection*{Acknowledgements}
We would like to thank Murray Shanahan for his helpful comments, critical feedback and insights regarding this work. We also thank Anna Hadjitofi for proof-reading and improving clarity throughout the writing of the paper.

\section*{Broader Impact}
% Authors should discuss both positive and negative outcomes, if any. For instance, authors should discuss a) who may benefit from this research, b) who may be put at disadvantage from this research, c) what are the consequences of failure of the system, and d) whether the task/method leverages biases in the data. If authors believe this is not applicable to them, authors can simply state this.
As it is with any machine learning model aimed at extracting patterns solely from data, learning invariants through soft unification is prone to being influenced by spurious correlations and biases that might be present in the data. There is no guarantee that even a clear, high accuracy invariant might correspond to a valid inference or casual relationship as discussed in Section~\ref{sec:analysis} with some mis-matching invariants presented in Appendix~\ref{apx:further_results}. As a result, if our approach succeeds in solving the task with an invariant, it does not mean that there is only pattern or in the case of failing to do so, a lack of patterns in the data. There has been recent work~\cite{invariantriskmin,invariantcausalpred} on tackling a different notion of invariance formed of features that are consistent (hence invariant) across different training dataset environments, to learn more robust predictors. Our method is instead targeted at research and researchers involved with combining cognitive aspects such as variable learning and assignment with neural networks under the umbrella of neuro-symbolic systems~\cite{neurosymbolic,neurosymsurvey}. A differentiable formulation of variables could accelerate the research of combining logic based symbolic systems with neural networks.
In summary, we regard this work as an experimental stepping stone towards better neuro-symbolic systems in the domain of artificial intelligence research.

\printbibliography

\clearpage
\appendix
\section{Model Details}\label{apx:model_details}
This appendix section describes each instance of the Unification Networks from Section~\ref{sec:inst_uni_nets} in more detail. In all cases, we select the hyper-parameters such as number of layers and embedding dimensions based on similar previous work.

\begin{table}[h]
\centering
\caption{A non-exhaustive list of symbols and notation used throughout the paper and their descriptions. Note that the notation in the Appendix sections might differ.}
\label{tab:notation_table}
\renewcommand{\arraystretch}{1.5}
\begin{tabular}{@{}cp{11cm}@{}}
\toprule
Notation & Description\\
\midrule
$G$ & Invariant example data point of which the variable symbols are learnt. It is derived from the term \textbf{g}round example that will be lifted.\\
$(C, Q, a)$ & A single data point consisting of a context, query and an answer.\\
$\psi$ & Variableness function that given a symbol outputs a value between 0 and 1.\\
$\sS$ & Set of all symbols appearing in a dataset or given example data point.\\
$I$ & Invariant that consists of an example $G$ and variableness function $\psi$.\\
$\sI$ & Set of invariants when multiple invariants are used.\\
\var & Variable symbol prefix, used to indicate symbols that are variables, i.e. have $\psi(s) \geq t$ above a certain threshold $t$.\\
\vard{bernhard} & Variable with default symbol bernhard, bernhard is the $s$ in $\psi(s) \geq t$.\\
$\phi$ & Features of a symbol, it could be any d-dimensional representation $\phi(s) \in \R^d$.\\
$\phi_U$ & Unifying features of a symbol, $\phi_U(s) \in \R^d$.\\
$K$ & A new example data point of which we want to the predict the answer.\\
$g$ & Soft unification function that computes the unified representation of $I$, defined in Algorithm~\ref{algo:uninet}.\\
$f$ & Upstream, potentially task specific, predictor network.\\
$\nll$ & The negative log-likelihood loss.\\
\bottomrule
\end{tabular}
\end{table}

\subsection{Unification MLP}
The input example is a sequence of symbols with a fixed length $l$, e.g. a sequence of digits \texttt{4234}. Given an embedded input $\phi(K) \in \R^{l \times d}$, the upstream MLP computes the output symbol based on the flattened representations $f(\phi(K)) = \operatorname{softmax}(\vh \mE^T)$ where $\vh \in \R^d$ is the output of the last layer and $\mE$ is the embedding matrix for the symbols. However, to compute the unifying features $\phi_U$, $g$ uses a bi-directional GRU~\cite{gru} running over $\phi(K)$ such that $\phi_U(K) = \Phi(K)\mW_U$ where $\Phi(K) \in \R^{l \times d}$ is the hidden state of the GRU at every symbol in $K$ and $\mW_U \in \R^{d \times d}$ is a learnable weight.

To model $f$ as a multi-layer perceptron, we take symbol embeddings of size $d=16$ and flatten sequences of length $l=4$ into an input vector of size $\phi(\vk) \in \R^{64}$. The MLP consists of 2 hidden layers with $\operatorname{tanh}$ non-linearity of sizes $2d$ and $d$ respectively and an output layer of size $d$. To process the query, we concatenate the one-hot encoding of the task id to $\phi(\vk)$ yielding a final input of size $64+4=68$. For unification features $\phi_U$, we use a bi-directional GRU with hidden size $d$ and the corresponding task id as the initial state. In this case we embed the task ids using another learnable embedding matrix. The hidden state at each symbol is taken with a linear transformation to give $\phi_U(s) = \mW_U\Phi(s)$ where $\Phi(s)$ is the hidden state of the bi-directional GRU. The variable assignment is then computed as an attention over the context sentence the according to Algorithm~\ref{algo:uninet}.

\subsection{Unification CNN}
Given a grid of embedded symbols $\phi(K) \in \R^{w \times h \times d}$ where $w$ is the width and $h$ the height, we use a convolutional neural network such that the final prediction is $f(\phi(K)) = \operatorname{softmax}((\mW\vh + \vb)\mE^T)$ where $\vh$ this time is the result of global max pooling and $\mW,\vb$ are learnable parameters. We also model $g$ using a \emph{separate} convolutional network with the same architecture as $f$ and set $\phi_U(K) = c_2(\operatorname{relu}(c_1(K))$ where $c_1, c_2$ are the convolutional layers. The grid is padded with 0s to obtain $w \times h \times d$ after each convolution such that every symbol has a unifying feature. This model conveys how soft unification can be adapted to the specifics of the domain, for example by using a convolution in a spatially structured input.

We take symbols embeddings of size $d=32$ to obtain an input grid $\phi(\mK) \in \R^{3 \times 3 \times 32}$. Similar to UMLP, for each symbol we append the task id as a one-hot vector to get an input of shape $3 \times 3 \times (32+4)$. Then $f$ consists of 2 convolutional layers with $d$ filters each, kernel size of 3 and stride 1. We use $\operatorname{relu}$ non-linearity in between the layers. We pad the grid with 2 columns and 2 rows to a $5 \times 5$ such that the output of the convolutions yield again a hidden output $\tH \in \R^{3 \times 3 \times d}$ of the same shape. As the final hidden output $h$, we take a global max pool to over $\tH$ to obtain $\vh \in \R^d$. Unification function $g$ is modelled identical to $f$ without the max pooling such that $\phi_U(\mK_{ij})=\tH'_{ij}$ where $\tH'$ is the hidden output of the convolutional layers.

\subsection{Unification RNN}
The input example is a variable length sequence of words such that $\phi(K) \in \R^{l \times d}$ where $l$ is the length of the sequence and $d$ is the embedding size. The upstream $f$ is modelled as an LSTM~\cite{lstm} which processes the input sequence to yield a final hidden state $\vh \in \R^d$ such that $\vh = \operatorname{LSTM}(\phi(K))$. To obtain the final prediction, we apply a linear transformation followed by the $\operatorname{sigmoid}$ non-linearity, $f(\phi(K)) = \sigma(\mW_f \vh + b_f)$ where $\mW_f \in \R^{1 \times d}$ and $b_f$ are learnable parameters. To compute the symbol features $\phi$, their unifying features $\phi_U$ and their variableness $\psi$, we use a linear transformation on the original word embedding of a symbol $\ve_s$ such that $\phi(s) = \mW\ve_s + \vb$, $\phi_U(s) = \mW\phi(s) + \vb$ and $\psi(s) = \sigma(\mW\phi(s) + b)$ where all $\mW$ and $\vb$ are \emph{distinct} learnable parameters.

In the sentiment analysis dataset, Section~\ref{sec:datasets}, we start with ConceptNet word embeddings~\cite{numberbatch} for each symbol $\ve_s \in \R^{300}$. The input words are then projected down to $d = 16$ dimensions to compute the required features above. We apply a dropout of 0.5 to $\vh$ and let the LSTM skip padding added to shorter sentences. The initial states of the LSTM are set as zero vectors.

\subsection{Unification Memory Networks}
The memory network $f$ uses the final hidden state of a bi-directional GRU (blue squares in Figure~\ref{fig:diag_unimemnn}) as the sentence representations to compute a context attention. At each iteration, we unify the words between the attended sentences using the same approach in UMLP with another bi-directional GRU (pink diamonds in Figure~\ref{fig:diag_unimemnn}) for unifying features $\phi_U(\text{bernhard}) = \mW_U\Phi(\text{bernhard})$. Following line~\ref{algoline:uninet_inv_value} in Algorithm~\ref{algo:uninet}, the new unified representation of the memory slot is computed and $f$ uses it to perform the next iteration. Concretely, $g$ produces an unification tensor $\tU \in \R^{M \times m \times N \times d}$ where $M$ and $m$ is the number of sentences and words in the invariant respectively, and $N$ is the number of sentences in the example such that after the context attentions are applied over $M$ and $N$, we obtain $\phi(\vk) \in \R^{m \times d}$ as the unified sentence with variables instantiated at that iteration. In other words, we compute all pairwise sentence unification and then use the context attentions from the invariant and the example to reduce the unification tensor $\tU$. Note that unlike in the UMLP case, the sentences can be of varying length. The prediction is then $\operatorname{softmax}(\mW\vh^J_I + \vb)$ where $\vh^J_I$ is the hidden state of the memory network running over the invariant after $J$ iterations. This setup, however, requires pre-training $f$ such that the context attentions match the correct pairs of sentences to unify.

Unlike previous architectures, with UMN we interleave $g$ into $f$. We use embedding sizes of $d=32$ and model $f$ with an iterative memory network. We take the final hidden state of a bi-directional GRU, with initial state $\mathbf{0}$, $\Phi_M$ to represent the sentences of the context $\mC$ and query $\vq$ in a $d$-dimensional vector $\mM_i = \Phi_M(\mC_i)$ and the query $\vm_q = \Phi_M(\vq)$. The initial state of the memory network is $\vh^0 = \vm_q$. At each iteration $j$:
\begin{align}
\mA^j_i &= \tanh(\mW\rho(\mM_i, \vh^j) + \vb) \label{eq:memnn}\\
\beta^j &= \text{softmax}(\mW\Phi_A(\mA^j) + \vb) \label{eq:ctx_att}
\end{align}
where $\Phi_A$ is another $d$-dimensional bi-directional GRU and $\rho(\vx,\vy) = [\vx;\vy;\vx \odot \vy;(\vx-\vy)^2]$ with $\odot$ the element-wise multiplication and $[;]$ the concatenation of vectors. Taking $\beta^j$ as the context attention, we obtain the next state of the memory network:
\begin{equation}
\vh^{j+1} = \sum_i \beta^j_i \tanh(\mW\rho(\mM_i, \vh^j) + \vb)
\end{equation}
and iterate $J$ many times in advance. The final prediction becomes $f(\mC,\vq) = \operatorname{softmax}(\mW\vh^J + \vb)$. All weight matrices $\mW$ and bias vectors $\vb$ are independent but are tied across iterations. The intuition is that $\mA^j$ captures some interaction between the current state and the memory slots. Then the bi-directional GRU is used to model the temporal relationship, e.g. if there are multiple sentences where Mary went, we would like to select the last location. Finally, another interaction is computed between each memory slot and the current state to obtain what the state would be if that memory slot was selected. Using $\beta^j$, the state is weighed based on which memory slot was selected at that iteration.

\newpage
\section{Generated Dataset Samples}\label{apx:dataset}

\begin{table}[h]
\centering
\caption{Sample context, query and answer triples from sequences and grid tasks.}
\label{tab:seq_grid_samples}
\renewcommand{\arraystretch}{2.5}
\begin{tabular}{ccccc}
\toprule
Dataset & Task & Context & Query & Answer\\
\midrule
%umlp_result/umlp_l4_s8_i4_e16_t0_f0.out
Sequence & i & 1488 & constant & 2\\
Sequence & ii & 6157 & head & 6\\
Sequence & iii & 1837 & tail & 7\\
Sequence & iv & 3563 & duplicate & 3\\
\midrule
%ucnn_result/ucnn_s8_i4_e32_t0_f0.out
Grid & i &
\setlength{\tabcolsep}{1mm}
\renewcommand{\arraystretch}{0.5}
\begin{tabular}{ccc}
 0 & 0 & 0\\
 0 & 2 & 2\\
 8 & 2 & 2
\end{tabular} & box & 2\\
Grid & ii &
\setlength{\tabcolsep}{1mm}
\renewcommand{\arraystretch}{0.5}
\begin{tabular}{ccc}
 4 & 0 & 0\\
 0 & 7 & 0\\
 8 & 0 & 1
\end{tabular} & head & 4\\
Grid & iii &
\setlength{\tabcolsep}{1mm}
\renewcommand{\arraystretch}{0.5}
\begin{tabular}{ccc}
 0 & 6 & 0\\
 1 & 7 & 2\\
 0 & 3 & 0
\end{tabular} & centre & 7\\
Grid & iv &
\setlength{\tabcolsep}{1mm}
\renewcommand{\arraystretch}{0.5}
\begin{tabular}{ccc}
 8 & 0 & 0\\
 5 & 6 & 0\\
 2 & 4 & 1
\end{tabular} & corner & 2\\
\bottomrule
\end{tabular}
\end{table}

%\begin{wraptable}{r}{0.4\linewidth}
\begin{table}[h]
\centering
\caption{Training sizes for randomly generated fixed length sequences and grid tasks with 8 unique symbols. The reason for Grid task (i) to be smaller is because there are at most 32 combinations of $2 \times 2$ boxes in a $3 \times 3$ grid with 8 unique symbols. The upper and lower bounds are taken from the 5-folds that are generated and might differ with each new random test split of the data.}
\label{tab:tsize}
\begin{tabular}{@{}rll@{}}
\toprule
Task & Sequences & Grid\\
\midrule
i   & $704.7\pm12.8$ & $25.6\pm1.8$\\
ii  & $709.4\pm13.8$ & $623.7\pm14.1$\\
iii & $709.7\pm14.0$ & $768.2\pm12.5$\\
iv  & $624.8\pm12.4$ & $795.2\pm10.3$\\
\bottomrule
\end{tabular}
\end{table}
%\end{wraptable}

\clearpage
\section{Training Details}\label{apx:training_details}
\subsection{Unification MLP \& CNN \& RNN}
\begin{figure}[h]
\centering
\includegraphics[width=0.9\linewidth]{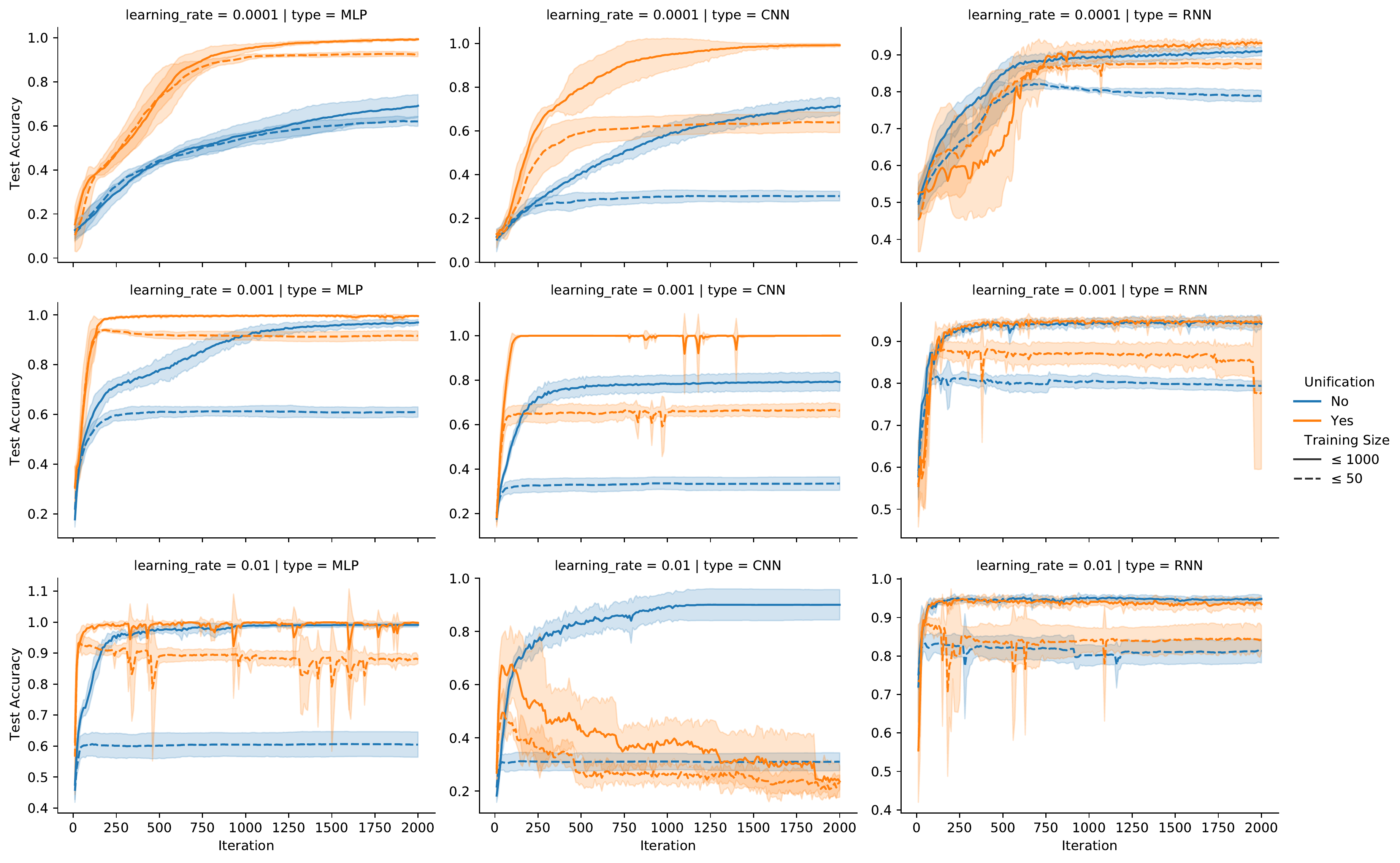}
\caption{Results of Unification MLP, CNN and RNN with different levels of learning rates. We observe that improvement is maintained except with 0.01 with which our approach degrades in training performance. With the more real-world dataset, we can see that our approach maintains a noisier (fluctuating) training curve.}
\label{fig:umlp_ucnn_learning_rates}
\end{figure}

Both unification models are trained on a 5-fold cross-validation over the generated datasets for 2000 iterations with a batch size of 64. In this context, each iteration is a single batch update. We don't use any weight decay and save the training and test accuracies every 10 iterations, as presented in Figure~\ref{fig:umlp_ucnn_acc}. We also present the training curves with different learning rates in Figure~\ref{fig:umlp_ucnn_learning_rates}.

\subsection{Unification Memory Networks}
We again use a batch size of 64 and pre-train $f$ for 40 epochs then $f$ together with $g$ for 260 epochs. We use epochs for UMN since the dataset sizes are fixed. To learn $g$ alongside $f$, we combine error signals from the unification of the invariant and the example. The objective function not only incorporates the negative log-likelihood $\nll$ of the answer but also the mean squared error between intermediate states $\vh^j_I$ and $\vh^j_K$ at each iteration as an auxiliary loss:
\begin{equation}\label{eq:objective}
J = \nll(f(K),a) + \lambda_I \left[ \nll(f \circ g(I,K),a) + \frac{1}{J}\sum^J_{j=0} (\vh^j_I - \vh^j_K)^2 + \tau\sum_{s \in \sS} \psi(s) \right]
\end{equation}
We pre-train by setting $\lambda_I=0$ for 40 epochs and then set $\lambda_I=1$. For strong supervision we also compute the negative log-likelihood $\nll$ for the context attention $\beta^j$, described in Appendix~\ref{apx:model_details}, at each iteration using the supporting facts of the tasks. We apply a dropout of 0.1 for all recurrent neural networks used and only for the bAbI dataset weight decay with 0.001 as the coefficient.

\subsection{Training Curves}

\begin{figure}[h]
\centering
\includegraphics[width=\linewidth]{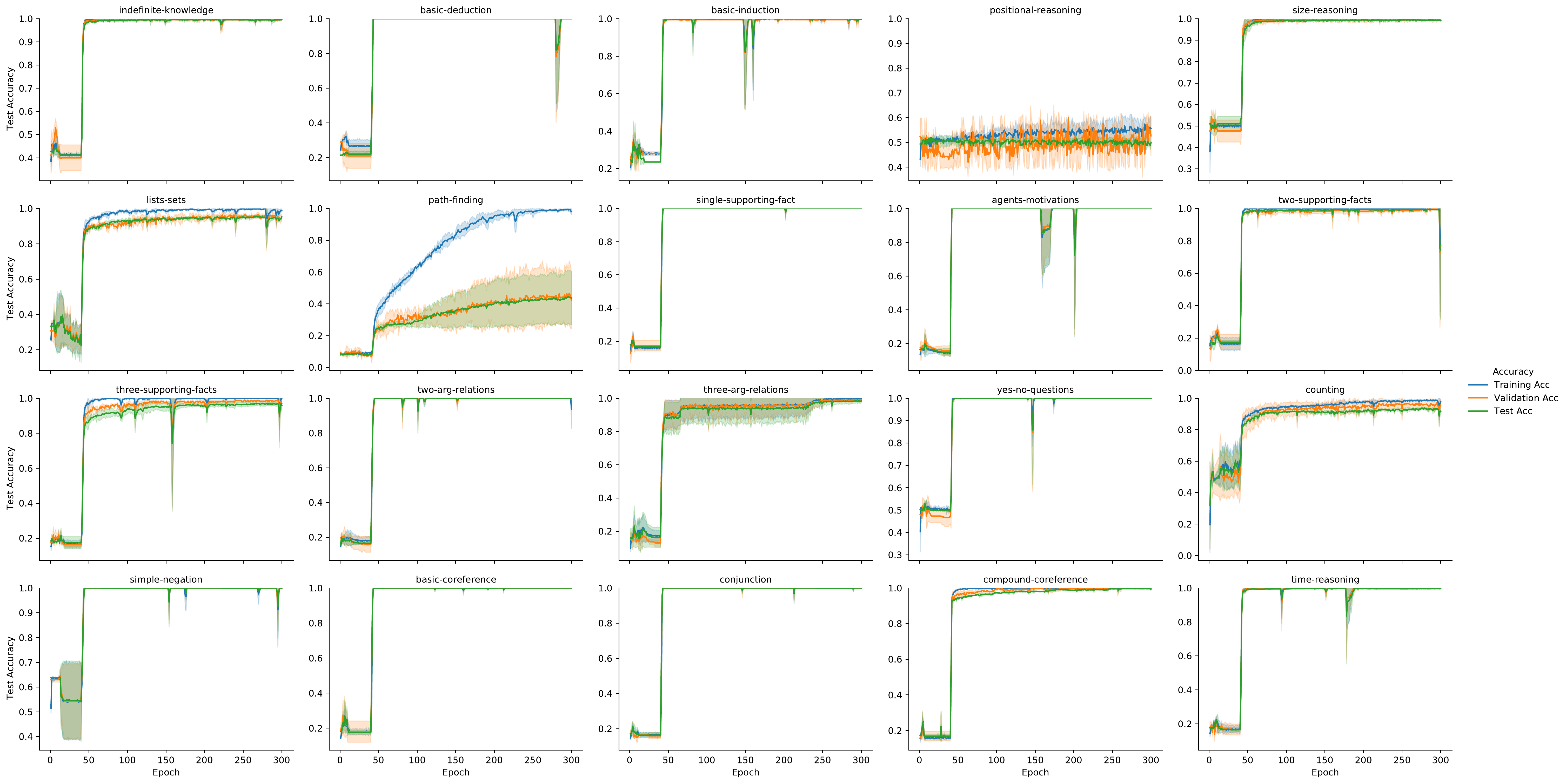}
\caption{Training curves for UMN on the bAbI dataset with strong supervision, 1 invariant and 1k training size. Note the initial 40 epochs are for pre-training the memory network and then unification is enabled which causes the sudden jump in the training curves. After unification is enabled, our approach learns to solve the tasks through soft unification within 5 epochs for most tasks. Tasks involving positional reasoning and path finding still prove difficult emphasising the dependency of our approach on the upstream memory network $f$ which struggles to solve or over-fits.}
\label{fig:babi_umn_training_curves}
\end{figure}

\begin{figure}[h]
\centering
\includegraphics[width=\linewidth]{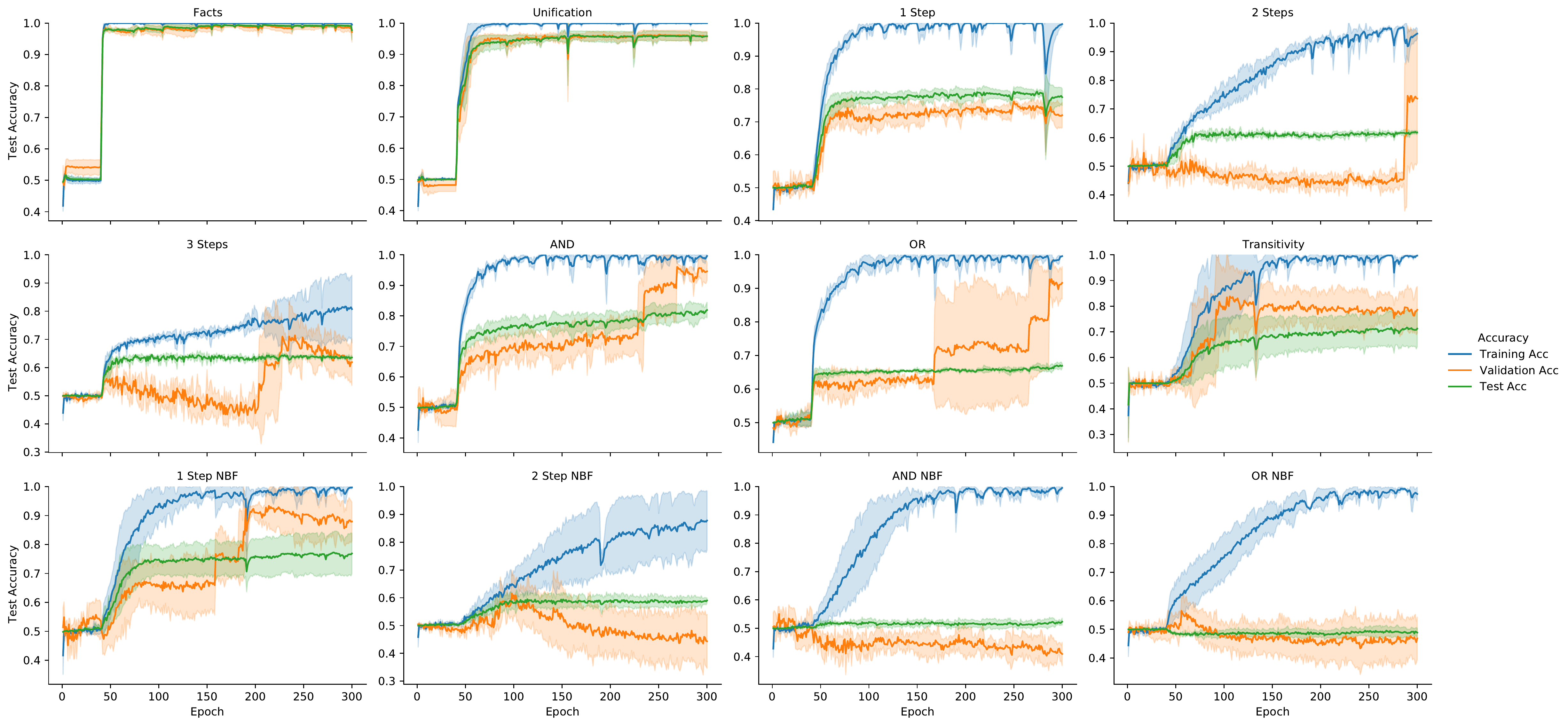}
\caption{Training curves for UMN on the logic dataset with strong supervision, 1 invariant and 2k training size. Note the initial 40 epochs are for pre-training the memory network and then unification is enabled. We observe that besides the very basic tasks that involve just Facts or Unification (these are single iteration tasks), our approach tends to over-fit.}
\label{fig:logic_umn_training_curves}
\end{figure}

\clearpage
\section{Further Results}\label{apx:further_results}

\begin{figure}[h]
\centering
\includegraphics[width=0.9\linewidth]{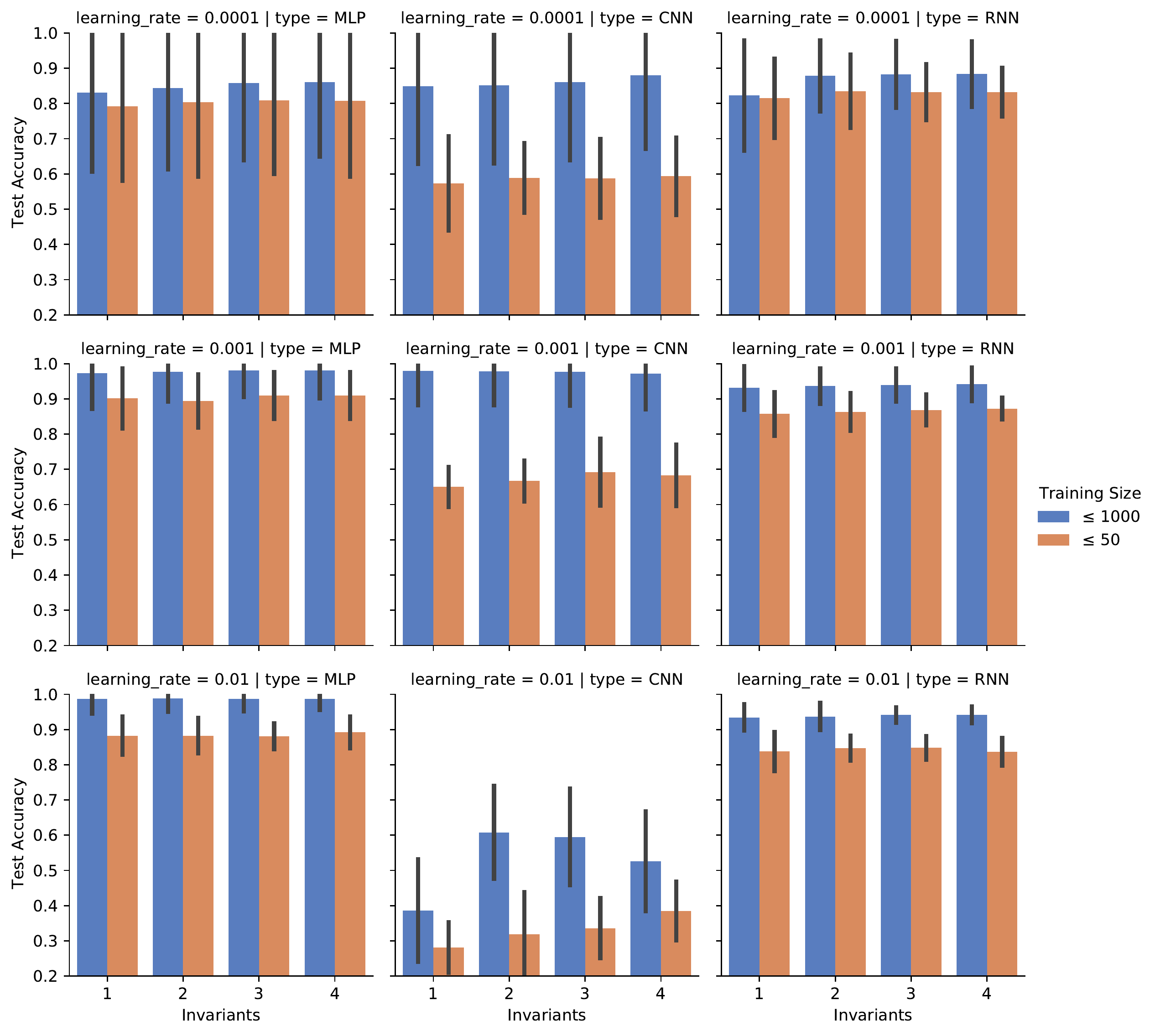}
\caption{Results of Unification MLP, CNN and RNN on increasing number of invariants with different learning rates. There is no impact on performance when more invariants per task are given. We speculate that the models converge to using 1 invariant because the datasets can be solved with a single invariant due to the regularisation applied on $\psi$ zeroing out unnecessary invariants.
%Results with multiple invariants are identical and the models learn to ignore the extra invariants as shown in Figure~\ref{fig:umlp_ucnn_invs} due to the regularisation applied on $\psi$ zeroing out unnecessary invariants.
This behaviour highlights the level of abstraction we can achieve through $\psi$. One might consider ``2 \var\ 7 \var'' and ``\var\ 8 4 \var'' to be different patterns or invariants but at a higher level of abstraction they can both represent the concept of a repeated symbol irrespective of the position of the repeating item. Thus, extra invariants can be ignored in favour of one that accounts for different examples through the learnable unifying features $\phi_U$, which might look for the notion of a \emph{repeated symbol} regardless of its position in the sequence. Given the capacity to learn unifying features and the pressure to use the least amount of variables, the models in these cases can optimise to learn and use a single invariant.
}
\label{fig:umlp_ucnn_invs}
\end{figure}

\begin{table}[h]
\small
\centering
\caption{Individual task error rates on bAbI tasks for Unification Memory Networks. Following previous work, these are the best error rates out of 3 runs and are obtained by taking the test accuracy at the epoch for which the validation accuracy is highest. For the variance across runs, you can refer to the sample training curves for each task in Figure~\ref{fig:babi_umn_training_curves}.}
\label{tab:babi_results}
\begin{tabular}{@{}rccccc@{}}
\toprule
Supervision & \multicolumn{2}{c|}{Weak} & \multicolumn{3}{c}{Strong}\\
\# Invs & 1 & 3 & 1 & 3 & 3\\
Training Size     & 1k & 1k & 1k & 1k & 50\\
\midrule
 1 &  0.0 &  0.0 &  0.0 &  0.0 &  1.1\\
 2 & 62.3 & 60.4 &  0.1 &  0.4 & 40.4\\
 3 & 58.8 & 63.7 &  1.2 &  1.3 & 52.1\\
 4 &  0.0 &  0.0 &  0.0 &  0.0 & 36.9\\
 5 &  1.9 &  1.6 &  0.5 &  1.6 & 29.7\\
 6 &  0.0 &  0.1 &  0.0 &  0.0 & 15.4\\
 7 & 20.5 & 22.3 &  6.4 &  7.7 & 22.4\\
 8 &  7.4 &  7.7 &  4.2 &  2.9 & 31.9\\
 9 &  0.3 &  0.0 &  0.0 &  0.0 & 20.6\\
10 &  0.1 &  0.5 &  0.2 &  0.3 & 26.5\\
11 &  0.0 &  0.0 &  0.0 &  0.0 & 21.1\\
12 &  0.0 &  0.0 &  0.0 &  0.0 & 23.5\\
13 &  0.4 &  4.7 &  0.0 &  0.2 &  5.6\\
14 & 15.3 & 17.4 &  0.1 &  0.1 & 57.3\\
15 & 17.8 &  0.0 &  0.0 &  0.0 &  0.0\\
16 & 52.7 & 53.3 &  0.0 &  0.0 & 45.4\\
17 & 39.9 & 49.3 & 49.5 & 48.4 & 45.8\\
18 &  7.2 &  7.9 &  0.3 &  0.8 & 10.9\\
19 & 90.4 & 90.7 & 38.9 & 67.8 & 86.2\\
20 &  0.0 &  0.0 &  0.0 &  0.0 &  1.8\\
\midrule
Mean & 18.8 & 19.0 & \textbf{5.1} & 6.6 & 28.7\\
Std & 27.7 & 28.0 & 13.6 & 18.0 & 21.8\\
\# $>5\%$ & 10 & 9 & 3 & 3 & 17\\
\bottomrule
\end{tabular}
\end{table}

\clearpage
\begin{table*}[h]
\small
\centering
\caption{Comparison of individual task error rates (\%) on the bAbI~\cite{babi} dataset of the best run. We preferred 1k results if a model had experiments published on both 1k and 10k for data efficiency. We present our approach (UMN) with a single invariant for weak and strong supervision, taken from Table~\ref{tab:babi_results}. References from left to right: \cite{memn2n} - \cite{gn2n} - \cite{entnet} - \cite{qrn} - Ours - \cite{dmnvisual} - \cite{dnc} - \cite{memnn} - Ours - \cite{dmn}}
\label{tab:babi_comp}
\begin{tabular}{@{}rcccccccccc@{}}
\toprule
Support & \multicolumn{7}{c|}{Weak} & \multicolumn{3}{c}{Strong} \\
Size & \multicolumn{5}{c|}{1k} & \multicolumn{2}{c|}{10k} & \multicolumn{2}{c}{1k} & 10k \\
Model & N2N & GN2N & EntNet & QRN & UMN & DMN+ & DNC & MemNN & UMN & DMN\\
%Ref & \cite{memn2n} & \cite{gn2n} & \cite{entnet} & \cite{qrn} & Ours & \cite{dmnvisual} & \cite{dnc} & \cite{memnn} & Ours & \cite{dmn}\\
\midrule
 1 &  0.0 &  0.0 &  0.7 &  0.0 &  0.0 &  0.0 &  0.0 &  0.0 &  0.0 &  0.0\\
 2 &  8.3 &  8.1 & 56.4 &  0.5 & 62.3 &  0.3 &  0.4 &  0.0 &  0.1 &  1.8\\
 3 & 40.3 & 38.8 & 69.7 &  1.2 & 58.8 &  1.1 &  1.8 &  0.0 &  1.2 &  4.8\\
 4 &  2.8 &  0.4 &  1.4 &  0.7 &  0.0 &  0.0 &  0.0 &  0.0 &  0.0 &  0.0\\
 5 & 13.1 &  1.0 &  4.6 &  1.2 &  1.9 &  0.5 &  0.8 &  2.0 &  0.5 &  0.7\\
 6 &  7.6 &  8.4 & 30.0 &  1.2 &  0.0 &  0.0 &  0.0 &  0.0 &  0.0 &  0.0\\
 7 & 17.3 & 17.8 & 22.3 &  9.4 & 20.5 &  2.4 &  0.6 & 15.0 &  6.4 &  3.1\\
 8 & 10.0 & 12.5 & 19.2 &  3.7 &  7.4 &  0.0 &  0.3 &  9.0 &  4.2 &  3.5\\
 9 & 13.2 & 10.7 & 31.5 &  0.0 &  0.3 &  0.0 &  0.2 &  0.0 &  0.0 &  0.0\\
10 & 15.1 & 16.5 & 15.6 &  0.0 &  0.1 &  0.0 &  0.2 &  2.0 &  0.2 &  2.5\\
11 &  0.9 &  0.0 &  8.0 &  0.0 &  0.0 &  0.0 &  0.0 &  0.0 &  0.0 &  0.1\\
12 &  0.2 &  0.0 &  0.8 &  0.0 &  0.0 &  0.0 &  0.0 &  0.0 &  0.0 &  0.0\\
13 &  0.4 &  0.0 &  9.0 &  0.3 &  0.4 &  0.0 &  0.1 &  0.0 &  0.0 &  0.2\\
14 &  1.7 &  1.2 & 62.9 &  3.8 & 15.3 &  0.2 &  0.4 &  1.0 &  0.1 &  0.0\\
15 &  0.0 &  0.0 & 57.8 &  0.0 & 17.8 &  0.0 &  0.0 &  0.0 &  0.0 &  0.0\\
16 &  1.3 &  0.1 & 53.2 & 53.4 & 52.7 & 45.3 & 55.1 &  0.0 &  0.0 &  0.6\\
17 & 51.0 & 41.7 & 46.4 & 51.8 & 39.9 &  4.2 & 12.0 & 35.0 & 49.5 & 40.6\\
18 & 11.1 &  9.2 &  8.8 &  8.8 &  7.2 &  2.1 &  0.8 &  5.0 &  0.3 &  4.7\\
19 & 82.8 & 88.5 & 90.4 & 90.7 & 90.4 &  0.0 &  3.9 & 64.0 & 38.9 & 65.5\\
20 &  0.0 &  0.0 &  2.6 &  0.3 &  0.0 &  0.0 &  0.0 &  0.0 &  0.0 &  0.0\\
\midrule
Mean & 13.9 & 12.7 & 29.6 & 11.3 & 18.8 & \textbf{2.8} & 3.8 & 6.7 & 5.1 & 6.4\\
\# $>5\%$ & 11 & 10 & 15 & 5 & 10 & 1 & 2 & 4 & 3 & 2\\
\bottomrule
\end{tabular}
\end{table*}

\begin{table*}[h]
\small
\centering
\caption{Individual task error rates (\%) of UMN against baseline IMA~\cite{deeplogic} with 2k training size. Following previous work in Table~\ref{tab:babi_agg_results}, we present the best task error rates across 3 runs. Each error rate is obtained by taking the test accuracy at the epoch when validation accuracy is highest. For the variance across runs, you can refer to the sample training curves for each task in Figure~\ref{fig:logic_umn_training_curves}.}
\label{tab:deeplogic_umn_ima_results}
\begin{tabular}{@{}rccccc|cc@{}}
\toprule
Model & \multicolumn{5}{c|}{UMN} & \multicolumn{2}{c}{IMA}\\
Training Size & \multicolumn{4}{c|}{2k} & 100 & \multicolumn{2}{c}{2k}\\
Supervision  & \multicolumn{2}{c|}{Weak} & \multicolumn{3}{c|}{Strong} & Weak & Strong\\
\# Invs & 1 & 3 & 1 & 3 & 3 & \multicolumn{2}{c}{-}\\
\midrule
Facts       &  1.6 &  0.9 &  0.2 &  0.3 & 37.0 &  2.8 &  3.8\\
Unification &  2.4 &  3.0 &  2.1 &  8.8 & 45.0 &  9.7 &  7.2\\
1 Step      & 48.9 & 45.2 & 21.1 & 18.7 & 49.2 & 42.6 & 36.3\\
2 Steps     & 49.7 & 49.1 & 37.6 & 36.9 & 48.4 & 49.5 & 36.7\\
3 Steps     & 48.9 & 49.9 & 35.4 & 36.9 & 46.6 & 48.3 & 35.9\\
AND         & 35.1 & 35.6 & 16.4 & 18.9 & 48.7 & 40.3 & 21.4\\
OR          & 37.2 & 38.1 & 32.3 & 32.9 & 45.9 & 38.4 & 25.3\\
Transitivity& 50.0 & 50.0 & 25.5 & 23.2 & 49.4 & 48.6 & 49.3\\
1 Step NBF  & 30.8 & 30.8 & 19.6 & 23.2 & 46.8 & 39.1 & 34.2\\
2 Steps NBF & 49.9 & 50.0 & 38.8 & 48.0 & 49.0 & 49.5 & 34.1\\
AND NBF     & 48.5 & 48.6 & 49.8 & 49.7 & 49.4 & 48.8 & 45.5\\
OR NBF      & 49.2 & 50.4 & 49.8 & 50.0 & 49.7 & 48.4 & 48.8\\
\midrule
Mean & 37.7 & 37.6 & \textbf{27.4} & 29.0 & 47.1 & 38.8 & 31.5\\
\# $>5\%$ & 10 & 10 & 10 & 11 & 12 & 11 & 11\\
\bottomrule
\end{tabular}
\end{table*}

\begin{table*}[h]
\small
\centering
\caption{Individual task error rates (\%) of UMN with extra configurations on the logical reasoning dataset as well as the reported results of baselines DMN and IMA from~\cite{deeplogic}. Note that the baselines results are trained on a different instance of the dataset that uses the same data generating script. When the logic programs are restricted to arity 1, the length of the predicates become fixed, e.g. $p(a)$, which simplifies soft unification by removing any ambiguity. In this case, the identity matrix would produce the correct value assignments for variables in our approach and as a result we see a clear improvement over arity 2 version of the dataset. Note that one cannot generate transitivity tasks with only arity 1 predicates. Generating more logic programs improves the overall performance of our apporach which fails to solve only 5 tasks with strong supervision and a single invariant.}
\label{tab:deeplogic_results}
\begin{tabular}{@{}rcccccccccc|cc@{}}
\toprule
Size     & \multicolumn{4}{c|}{2k} & \multicolumn{5}{c}{20k} & 100 & \multicolumn{2}{c}{40k}\\
Support  & \multicolumn{2}{c|}{Weak} & \multicolumn{2}{c}{Strong} & \multicolumn{2}{c|}{Weak} & \multicolumn{4}{c|}{Strong} & \multicolumn{2}{c}{Weak}\\
Arity    & 1 & 2 & 1 & 2 & 1 & 2 & 1 & 2 & 2 & 2 & 2 &2\\
\# Invs / Model & 1 & 3 & 1 & 3 & 1 & 3 & 1 & 1 & 3 & 3 & DMN & IMA\\
\midrule
Facts       &  1.2 &  0.9 &  0.0 &  0.4 &  0.0 &  0.0 &  0.0 &  0.0 &  0.0 & 33.5 &  0.0 &  0.0\\
Unification &  0.0 & 10.3 &  0.0 & 10.8 &  0.0 &  0.0 &  0.0 &  0.0 &  0.0 & 41.3 & 13.0 & 10.0\\
1 Step      & 50.3 & 49.8 &  4.4 & 20.0 &  1.2 & 27.8 &  0.1 &  1.3 &  5.7 & 50.2 & 26.0 &  2.0\\
2 Steps     & 47.5 & 50.0 &  5.7 & 35.0 & 37.2 & 47.8 &  0.0 & 29.7 & 28.7 & 49.9 & 33.0 &  5.0\\
3 Steps     & 47.6 & 49.2 & 10.4 & 38.7 & 39.6 & 45.6 &  0.0 & 26.0 & 26.1 & 48.3 & 23.0 &  6.0\\
AND         & 31.3 & 37.4 & 10.7 & 16.4 & 29.8 & 29.0 &  0.2 &  0.4 &  1.2 & 50.0 & 20.0 &  5.0\\
OR          & 25.2 & 38.1 & 21.0 & 35.0 & 20.5 & 30.2 &  4.4 & 20.6 & 17.4 & 47.6 & 13.0 &  3.0\\
Transitivity&      & 50.0 &      & 26.6 &      & 39.6 &      &  5.0 &  6.0 & 49.2 & 50.0 & 50.0\\
1 Step NBF  & 46.4 & 38.7 &  3.8 & 28.8 &  1.1 & 21.6 &  0.1 &  1.1 &  8.0 & 47.6 & 21.0 &  2.0\\
2 Steps NBF & 48.5 & 48.9 &  7.7 & 39.6 & 30.4 & 48.2 &  0.1 & 33.4 & 28.7 & 50.3 & 15.0 &  4.0\\
AND NBF     & 51.0 & 50.1 & 43.1 & 48.6 & 29.4 & 44.2 &  0.1 &  1.3 & 40.1 & 49.5 & 16.0 &  8.0\\
OR NBF      & 51.4 & 48.4 & 50.8 & 47.3 & 47.6 & 47.8 & 21.3 & 27.6 & 30.5 & 47.3 & 25.0 & 14.0\\
\midrule
Mean & 36.4 & 39.3 & 14.3 & 28.9 & 21.5 & 31.8 & \textbf{2.4} & 12.2 & 16.0 & 47.1 & 21.2 & 9.1\\
Std & 18.7 & 15.9 & 16.4 & 14.1 & 17.1 & 16.7 & 6.1 & 13.2 & 13.6 & 4.7 & 12.3 & 13.4\\
\# $>5\%$ & 9 & 11 & 7 & 11 & 7 & 10 & 1 & 5 & 9 & 12 & 11 & 5\\
\bottomrule
\end{tabular}
\end{table*}

\begin{figure}[h]
\centering
\renewcommand{\arraystretch}{1.3}
\begin{tabular}{l}
  \vard{sandra} went back to the \vard{bathroom}\\
  \hline
  is \vard{sandra} in the \vard{bathroom}\\
  \hline
  yes
\end{tabular}
\caption{bAbI task 6, yes or no questions. The invariant captures the varying components of the story which are the person and the location they have been, similar to task 1 of the bAbI dataset. However, the invariant captures nothing about how this task is actually solved by the upstream network $f$, i.e. how $f$ uses the interpolated unified representation.}
\label{fig:qa06_r1_s1_t0_s0_task6}
\end{figure}

\begin{figure}[h]
\centering
\renewcommand{\arraystretch}{1.3}
\begin{tabular}{rl}
 \vard{m} ( \vard{e} ) & $\vdash$ \vard{m} ( \vard{e} )\\
 \vard{a} ( \vard{w} , \vard{e} ) & $\vdash$ \vard{a} ( \vard{w} , \vard{e} )\\
 \vard{m} ( \texttt{T} ) & $\vdash$ \vard{m} ( c )\\
 \vard{x} ( \texttt{A} ) $\leftarrow$ \textit{not} \vard{q} ( \texttt{A} ) & $\vdash$ \vard{x} ( \vard{z} )
\end{tabular}
\caption{Invariants learned on tasks 1, 2 and 11 with arity 1 and 2 from the logical reasoning dataset. Last invariant on task 11 lifts the example around the negation by failure, denoted as \textit{not}, capturing its semantics. We present the context on the left and the query on the right, $C \vdash q$ means $f(C,q)=1$ as described in Section~\ref{sec:datasets}.}
\label{fig:rule_deeplogic_1_2}
\end{figure}

\begin{figure}[h]
\centering
\begin{subfigure}[t]{\linewidth}
\centering
\setlength{\tabcolsep}{0.2em}
\begin{tabular}{rl}
%for file in umlp_result/umlp_l4_s8_i1_e16_t0_f*.out; do sed -n 3,18p $file; done
3 \vard{4} 7 \vard{8} & head\\
7 4 \vard{2} \vard{6} & head\\
\vard{4} 3 1 \vard{5} & tail\\
\vard{3} \vard{3} 5 \vard{6} & duplicate
\end{tabular}
\caption{Invariants with extra variables learned with UMLP.}
\label{fig:fail_rule_seq_tasks}
\end{subfigure}
\begin{subfigure}[t]{\linewidth}
\centering
\begin{tabular}{ccc}
%for file in ucnn_result/ucnn_s8_i1_e32_t0_f*.out; do sed -n 191,200p $file; done
\setlength{\tabcolsep}{0.2em}
\begin{tabular}{@{}rrr@{}}
 \vard{4} & 0 & 0\\
 0 & 1 & 0\\
 0 & 0 & \vard{3}
\end{tabular} &
%for file in ucnn_result/ucnn_s8_i1_e32_t0_f*.out; do sed -n 313,322p $file; done
\setlength{\tabcolsep}{0.2em}
\begin{tabular}{@{}rrr@{}}
 0 & \vard{4} & 0\\
 1 & \vard{6} & 8\\
 0 & 2 & 0
\end{tabular} &
%for file in ucnn_result/ucnn_s8_i1_e32_t0_f*.out; do sed -n 451,460p $file; done
\setlength{\tabcolsep}{0.2em}
\begin{tabular}{@{}rrr@{}}
 6 & 4 & \vard{3}\\
 0 & \vard{2} & 8\\
 0 & 0 & 7
\end{tabular}\\
head & centre & corner
\end{tabular}
\caption{Mismatching invariants learned with UCNN.}
\label{fig:fail_rule_grid_tasks}
\end{subfigure}
\caption{Invariants learned that do not match the data generating distribution from UMLP and UCNN using $\le 1000$ examples to train. In these instances the unification still binds to the the correct symbols in order to predict the desired answer, i.e. quantitatively we get the same results. With these invariants, either the upstream network $f$ or the unifying features $\phi_U$ compensate for the extra or mis-matching variables.}
%\label{fig:qaxx_r1_s1_t0_s0}
%\label{fig:dlxx_arx_10k_r1_s1_t0_s0}
\label{fig:fail_invariants}
\end{figure}

\begin{figure}[h]
\centering
\begin{subfigure}[t]{0.32\textwidth}
\centering
\includegraphics[width=0.87\textwidth]{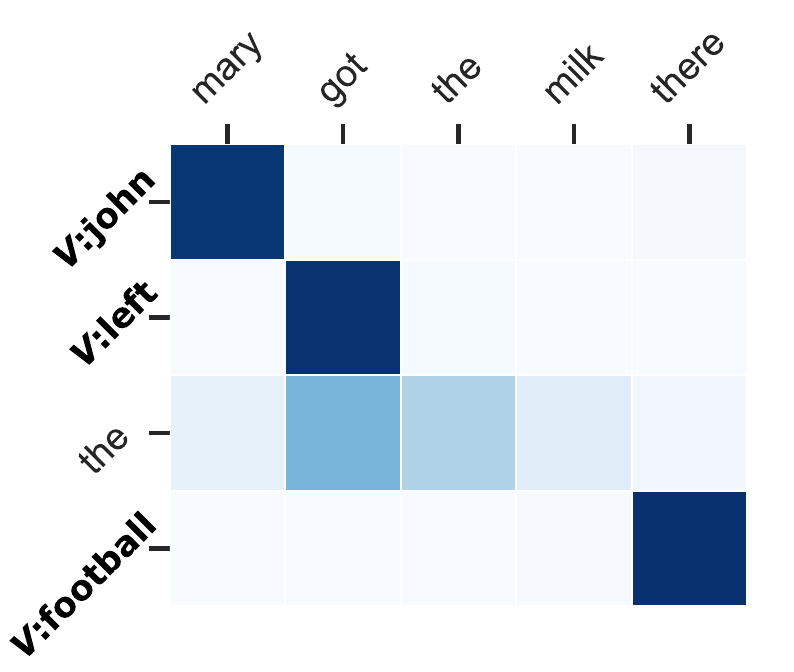}
\caption{bAbI task 2. When a variable is unused in the next iteration, e.g. \vard{football}, it unifies with random tokens often biased by position.}
\label{fig:uni_babi_02}
\end{subfigure}
~
\begin{subfigure}[t]{0.32\textwidth}
\centering
\includegraphics[width=0.64\textwidth]{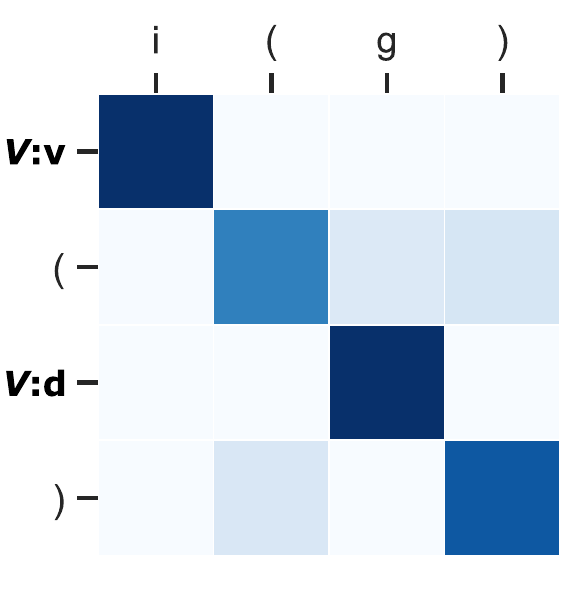}
\caption{Logical reasoning task 1. A one-to-one alignment is created between predicates and constants.}
\label{fig:uni_dl_01}
\end{subfigure}
~
\begin{subfigure}[t]{0.32\textwidth}
\centering
\includegraphics[width=0.9\textwidth]{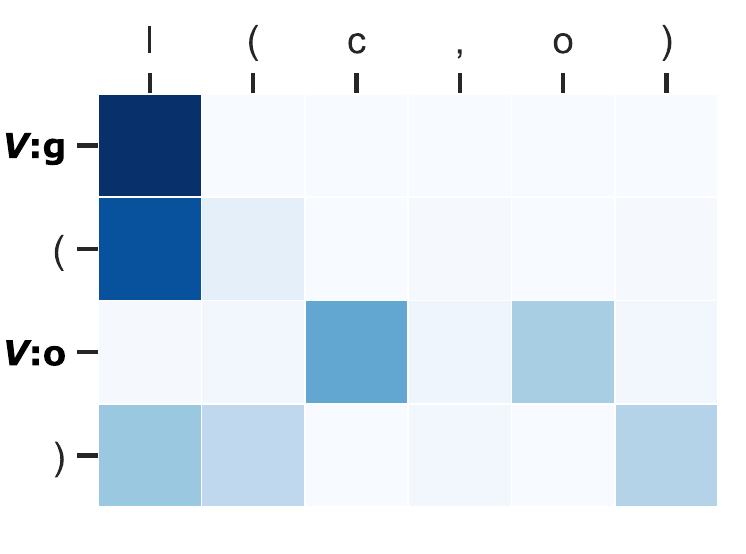}
\caption{Logical reasoning task 3. Arity 1 atom forced to bind with arity 2 creates a one-to-many mapping.}
\label{fig:uni_dl_03}
\end{subfigure}
\caption{Further attention maps from line~\ref{algoline:uninet_uni_att} in Algorithm~\ref{algo:uninet}, darker cells indicate higher attention values.}
\label{fig:uni_more}
\end{figure}

\end{document}